\documentclass[final]{cvpr}

\usepackage{amssymb}

\usepackage{times}
\usepackage{epsfig}
\usepackage{graphicx}
\usepackage{amsmath}
\usepackage{amssymb}
\usepackage{bm}
\usepackage{tabularx}
\usepackage{multirow}
\usepackage{booktabs}
\usepackage{caption}
\usepackage{threeparttable}

\usepackage{xcolor}
\usepackage[export]{adjustbox}
\usepackage{graphbox}

\usepackage{breqn}
\makeatletter
\newcommand{\subalign}[1]{%
  \vcenter{%
    \Let@ \restore@math@cr \default@tag
    \baselineskip\fontdimen10 \scriptfont\tw@
    \advance\baselineskip\fontdimen12 \scriptfont\tw@
    \lineskip\thr@@\fontdimen8 \scriptfont\thr@@
    \lineskiplimit\lineskip
    \ialign{\hfil$\m@th\scriptstyle##$&$\m@th\scriptstyle{}##$\hfil\crcr
      #1\crcr
    }%
  }%
}
\makeatother

\usepackage[pagebackref=true,breaklinks=true,colorlinks,bookmarks=false]{hyperref}
\usepackage{cleveref}
\Crefname{equation}{Eq.}{Eqs.}
\Crefname{figure}{Fig.}{Figs.}
\Crefname{tabular}{Tab.}{Tabs.}
\Crefname{table}{Tab.}{Tabs.}
\newcommand\blfootnote[1]{%
  \begingroup
  \renewcommand\thefootnote{}\footnote{#1}%
  \addtocounter{footnote}{-1}%
  \endgroup
}

\def\cvprPaperID{4721} 
\def\confYear{CVPR 2021}

\begin{document}

\title{Smoothing the Disentangled Latent Style Space for \\ Unsupervised Image-to-Image Translation}

\author{
  Yahui Liu\textsuperscript{1,3},
  Enver Sangineto\textsuperscript{1},
  Yajing Chen\textsuperscript{2},
  Linchao Bao\textsuperscript{2}, 
  Haoxian Zhang\textsuperscript{2}, \\
  Nicu Sebe\textsuperscript{1},
  Bruno Lepri\textsuperscript{3},
  Wei Wang\textsuperscript{1}\textsuperscript{*},
  Marco De Nadai\textsuperscript{3}\thanks{Both authors contributed equally to this
work.} \vspace{0.2cm}
\\
  \textsuperscript{1}University of Trento, Italy \quad
  \textsuperscript{2}Tencent AI Lab, China \quad
  \textsuperscript{3}Fondazione Bruno Kessler, Italy\\
}

\twocolumn[{%
\renewcommand\twocolumn[1][]{#1}%
\vspace{-3em}
\maketitle
\vspace{-3em}
\begin{center}
    \centering
    \includegraphics[width=0.95\linewidth]{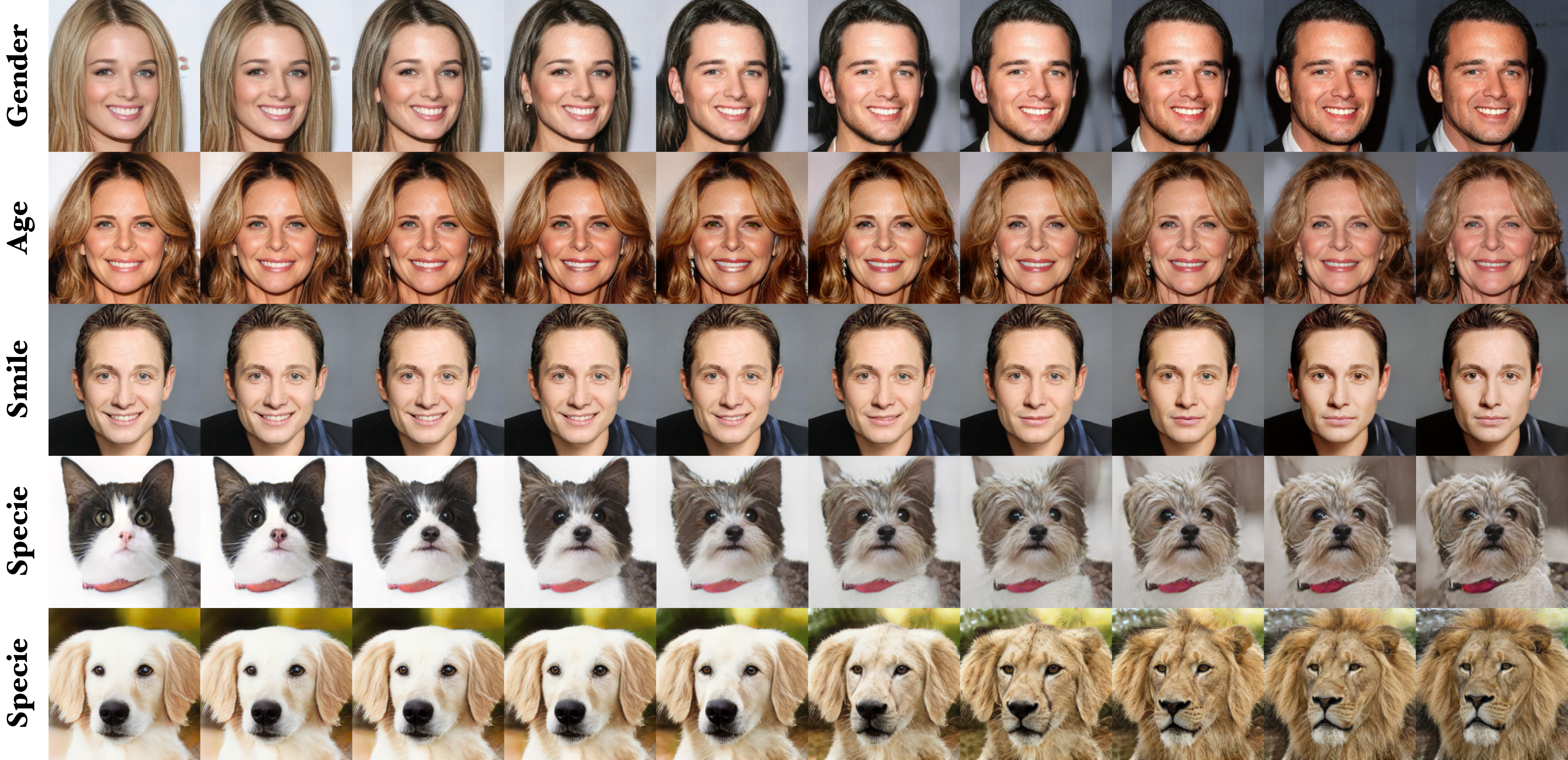}
    \vspace{-0.6em}
    \captionof{figure}{
    Our method generates smooth interpolations within and across domains in various image-to-image translation tasks. Here, we show gender, age and smile translations from CelebA-HQ~\cite{karras2017progressive} and animal translations from AFHQ~\cite{choi2019stargan}.
    }
    \label{fig:teaser}
\end{center}%
}]

\thispagestyle{empty}
\pagestyle{empty}

\begin{abstract}
\vspace{-1.5em}

Image-to-Image (I2I) multi-domain translation models are usually evaluated also using the quality of their semantic interpolation results. However, state-of-the-art models frequently show abrupt changes in the image appearance during interpolation, and usually perform poorly in interpolations across domains. In this paper, we propose a new training protocol based on three specific losses which help a translation network to learn a smooth and disentangled latent style space in which: 1) Both intra- and inter-domain interpolations correspond to gradual changes in the generated images and 2) The content of the source image is better preserved during the translation. Moreover, we propose a novel evaluation metric to properly measure the smoothness of latent style space  of I2I translation models. The proposed method can be plugged in existing translation approaches, and our extensive experiments on different datasets show that it can significantly boost the quality of the generated images and the graduality of the interpolations.

\end{abstract}

\begin{figure*}[t]	
	\renewcommand{\tabcolsep}{1pt}
	\newcommand{\sizea}{0.94\linewidth}
	\centering
	\footnotesize
	\begin{tabular}{c}
	    \includegraphics[width=\sizea]{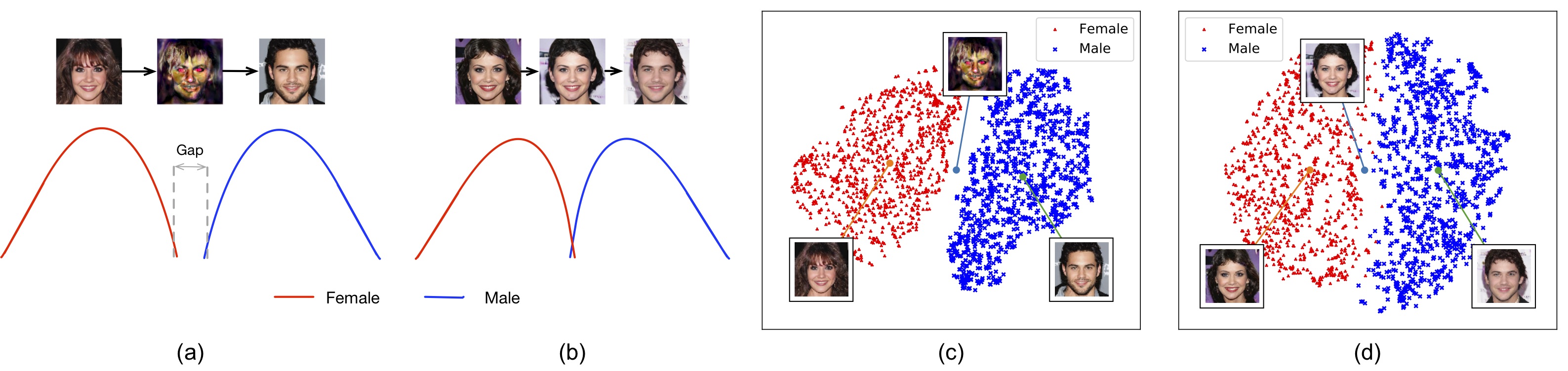}
	\end{tabular}
	\vspace{-1.2em}
	\caption{An illustration of the relation between smoothness and disentanglement of the style space. (a) Two well-separated distributions with a large  margin in between. The intermediate area  can lead to the generation of  artifacts because it has not been sufficiently explored during training. (b) When the margin is reduced, the corresponding image appearance changes are smoother. (c) A t-SNE visualization of randomly sampled style codes using StarGAN v2~\cite{choi2019stargan}, which shows a disentangled  style space but also that the inter-domain area generates images with artifacts.
	(d) The same visualization shows that, using  our  method, despite the disentanglement is preserved, the inter-domain area  generates realistic images.}
	\label{Fig:interpolation-intro}
\end{figure*}

\section{Introduction}
\label{sec:introduction}
\blfootnote{* These two authors contributed equally to this work. Correspondence to: wei.wang@unitn.it and work@marcodena.it.}
Translating images from one domain to another is a challenging image manipulation task that has recently drawn increasing attention in the computer vision community~\cite{choi2018stargan,choi2019stargan,huang2018multimodal,isola2017image,lee2019drit++,10.1145/3394171.3413505,PAMIStephane,zhu2017unpaired}. A 
``domain'' refers to a set of images sharing some distinctive visual pattern,  usually called 
``style'' (e.g., the gender or the hair color in face datasets)~\cite{choi2019stargan, huang2018multimodal,zhu2017unpaired}. 
The Image-to-Image (I2I) translation task aims to change the domain-specific aspects of an image while preserving its  ``content'' (e.g., the identity of a person or the image background)~\cite{huang2018multimodal}.
Since paired data (e.g., images of the same person with different gender) are usually not available, an important aspect of I2I translation models is the unsupervised training~\cite{zhu2017unpaired}.
Moreover, it is usually desirable to synthesize the multiple appearances {\em modes} within the same style domain, in such a way to be able to generate \emph{diverse} images for the same input image.

Recent work addresses the I2I translation using multiple domains~\cite{choi2018stargan,lee2019drit++,choi2019stargan} and generating multi-modal outputs~\cite{lee2019drit++, choi2019stargan}. 
These Multi-domain and Multi-modal Unsupervised Image-to-Image Translation (MMUIT) models are commonly evaluated based on the quality and the diversity of the generated images, including 
the results obtained by interpolating between two endpoints in their latent representations
(e.g., see \Cref{fig:teaser}).
However, interpolations are usually computed using only points belonging to the same domain, and most of the state-of-the-art MMUIT methods are inclined to produce artifacts or unrealistic images when tested using across-domain interpolations.
This is shown in \Cref{Fig:interpolation-intro} (c), where, using the state-of-the-art StarGAN v2~\cite{choi2019stargan}, the inter-domain area in the style space frequently generates artifacts.
Another common and related problem is the lack of graduality in both  intra and  inter domain interpolations, i.e., the generation of abrupt appearance changes 
corresponding to two close points in the latent space.

In this paper, we address the problem of learning a smoothed and disentangled style space for MMUIT models, which can be used for gradual and realistic image interpolations within and across domains. With ``disentangled'' we mean that the representations of different domains are well separated and clustered (\Cref{Fig:interpolation-intro}), so that intra-domain interpolations correspond to only intra-domain images. With ``smoothed'' we mean that the semantics of the style space changes gradually and these changes correspond to small changes in the human perceptual similarity.

The main idea of our proposal is based on the hypothesis that the interpolation problems are related to the exploration of latent space areas which correspond to sparse training data. We again refer to \Cref{Fig:interpolation-intro} to illustrate the intuition behind this observation. Many MMUIT methods use adversarial discriminators to separate the distributions of different domains~\cite{choi2019stargan}. However, a side-effect of this disentanglement process is that some areas of the latent space do not correspond to real data observed during training. Consequently, when interpolating in those areas, the decoding process may lead to generating unrealistic images. 
We propose to solve this problem jointly using a  triplet loss~\cite{schroff2015facenet,balntas2016learning} and a simplified version of the Kullback-Leibler (KL) divergence regularization~\cite{kingma2013auto}. The former separates the domains using a small {\em margin} on their relative distance, while the latter encourages the style codes to lie in a compact space.
The proposed simplified KL regularization does not involve the estimation of parametric distributions~\cite{kingma2013auto} and it can be easily plugged in Generative Adversarial Networks (GANs) \cite{choi2019stargan,baek2020tunit}.
On the other hand, differently from  adversarial discrimination,
the triplet-loss 
 margin can {\em control} the inter-domain distances  and help to preserve the domain disentanglement  in the compact space, 
Finally,  we also encourage the content \emph{preservation}  during the translation using a perceptual-distance based loss.
\Cref{fig:teaser} 
shows some interpolation results obtained using our method. In Sec.~\ref{sec:experiments} we qualitatively and quantitatively evaluate our approach and we show that it can be plugged in different existing MMUIT methods improving their results.
The last contribution of this paper concerns the proposal of the Perceptual Smoothness (PS) metric based on the perceptual similarity of the interpolated images, to quantitatively evaluate the style smoothness in MMUIT models.

The \textbf{contributions} of this paper can be  summarized as follows. First, we propose a new training strategy based on three specific losses 
which improve the interpolation smoothness and the content preservation of different MMUIT models.  Second, we propose a novel metric to fill-in the gap of previous MMUIT evaluation protocols and quantitatively measure the smoothness of the style space.

\section{Related Work}
\label{sec:related_work}

\noindent \textbf{Unsupervised Domain Translation}. Translating images from one domain to another without paired-image supervision is a  challenging task. Different constraints have been proposed to narrow down the space of feasible mappings between images.
Taigman \emph{et al.}~\cite{taigmanPW17} minimize the feature-level distance between the generated \ and the source image. Liu \emph{et al.}~\cite{liu2017unsupervised} create a shared latent space between the domains, which encourages different images to be mapped into the same space. CycleGAN~\cite{zhu2017unpaired} uses a cycle consistency loss in which the generated image is translated back to the original domain (an approach proved to be pivotal in the field~\cite{kim2017learning,alami2018unsupervised,mo2018instanceaware}). 
However, all these approaches are limited to one-to-one domain translations, thus requiring $m(m{-}1)$ trained models for  translations with $m$ domains.
StarGAN~\cite{choi2018stargan} was the first  single-model for \textit{multi-domain} translation settings. The generation process is conditioned by a target domain label,  input to the generator, and by a domain classifier in the discriminator. However, the I2I translation of StarGAN is  deterministic, since, for a given source image and target domain, only one target image can be generated (no multi-modality).

\noindent \textbf{Multi-modal and Multi-domain Translation}. 
After the pioneering works in supervised and one-to-one image translations~\cite{zhu2017toward, huang2018multimodal, MSGAN}, the recent literature is mainly focused in multiple-domains and multi-modal translations. 
Both DRIT++~\cite{lee2019drit++} and SMIT~\cite{romero2019smit}  use a noise input vector and a domain label to increase the output diversity.
StarGAN v2~\cite{choi2019stargan} relies on a multitask discriminator~\cite{liu2019few} to model multiple domains, a noise-to-style mapping network, and a diversity sensitive loss~\cite{MSGAN} to explore the image space better. However, qualitative results show changes of subtle ``content'' details (e.g., the color of the eyes, the shape of the chin or the background) while translating the image with respect to the style (e.g., the hair colour or the gender).

Although MMUIT models do not require any image-level supervision, they still require set-level supervision (i.e. domain labels for each image). Very recently, TUNIT~\cite{baek2020tunit} proposed a ``truly unsupervised" task where the network does not need any supervision. TUNIT learns the set-level characteristics of the images (i.e., the domains), and then it learns to map the images to all the domains. We will empirically show that our method can be used with both StarGAN v2 and TUNIT, and significantly improve the interpolation smoothness with both models.

\noindent \textbf{Latent-space interpolations.} There is a quickly growing interest in the recent I2I translation literature with respect to  latent space interpolations as a byproduct of the translation task. However, most previous works are only qualitatively evaluated, they use only intra-domain interpolations~\cite{lee2018diverse,lee2019drit++,romero2019smit}, or they require specific architectural choices.
For example, DLOW~\cite{gong2019dlow} is a one-to-one domain translation, and RelGAN~\cite{wu2019relgan} uses a linear interpolation loss at training time, but it is not multi-modal. 
In StarGAN v2~\cite{choi2019stargan}, the style codes of different domains are very well disentangled, but the inter-domain interpolations show low-quality results  (e.g., see \Cref{Fig:interpolation-intro}).
HomoGAN~\cite{chen2019homomorphic}  learns an explicit linear interpolator between images, but the generated images have very limited diversity.

Interestingly, image interpolations are not limited to the I2I translation field. The problem is well studied in Auto-Encoders~\cite{kingma2013auto,bojanowski2017optimizing, berthelot2018understanding} and in GANs~\cite{arvanitidis2017latent, karras2019style, karras2020analyzing}, where the  image is encoded into the latent space without an explicit separation between content and style. For example, StyleGAN~\cite{karras2019style} and StyleGANv2~\cite{karras2020analyzing} show high-quality interpolations of the latent space, where the latter  has  been further  studied to identify the emerging semantics (e.g. linear subspaces) without retraining the network~\cite{shen2020interpreting,jahanian2019steerability,zhu2020domain}. Richardson \emph{et al.}~\cite{richardson2020encoding}  propose to find the latent code of a real image in the pre-trained StyleGAN space. This two-stage inversion problem allows multi-modal one-to-one domain mappings and interpolations.
However, these methods are not designed to keep the source-image content while changing the domain-specific appearance. Thus, they are not suitable for a typical MMUIT task.

\section{Problem Formulation and Notation}
\label{sec:problem-formulation}

    Let $\pmb{\mathcal{X}} = \bigcup_{k=1}^m \pmb{\mathcal{X}}_k$ be the image set composed of $m$ disjoint domains
    ($\pmb{\mathcal{X}}_i \cap \pmb{\mathcal{X}}_j = \emptyset, i \neq j$), where each domain  $\pmb{\mathcal{X}}_k$ contains images sharing the same style. 
     The goal of a multi-domain I2I translation model is to learn a single functional $G(i,j) = \pmb{\mathcal{X}}_i \to \pmb{\mathcal{X}}_j$ for all possible $i,j \in \{1, 2, \cdots, m\}$.
     The domain identity  can be represented either using a discrete domain label (e.g., $i$) or by means of
     a style code $\pmb{s}$, where $\pmb{s} \in \pmb{\mathcal{S}}$ is a continuous vector and the set $\pmb{\mathcal{S}}$ of all the styles 
     may be either shared among all the domains or it can be partitioned in different domain-specific subsets (i.e., $\pmb{\mathcal{S}}=\{\pmb{\mathcal{S}}_1, \cdots, \pmb{\mathcal{S}}_m\}$). In our case, we use the second solution
     and  we denote with $\hat{\pmb{x}} = G(\pmb{x},\pmb{s})$ the translation operation, where $\pmb{x} \in \pmb{\mathcal{X}}_i$ is the {\em source} image (and its domain implicitly indicates the source domain $i$), $\pmb{s} \in \pmb{\mathcal{S}}_j$
     is the {\em target} style code and $\hat{\pmb{x}} \in \pmb{\mathcal{X}}_j$ is the generated image.

The MMUIT task is an extension of the above description in which:
\begin{enumerate}\vspace{-0.3\baselineskip}
    \item[a.] \textit{Training is unsupervised.} This is crucial  when collecting paired images is time consuming or  impossible.\vspace{-0.4\baselineskip}
    \item[b.] 
    \textit{The source content is preserved.} A translated image $\hat{\pmb{x}} = G(\pmb{x},\pmb{s})$ should preserve domain-invariant characteristics (commonly called ``content'') and change only the domain-specific properties of the source image $\pmb{x}$. For example, in $\text{male} \leftrightarrow \text{female}$ translations, $\hat{\pmb{x}}$ should keep the pose and the identity of  $\pmb{x}$, while changing 
    other aspects to look like a female or a male.\vspace{-0.4\baselineskip}
    \item[c.] 
    \textit{The output is multi-modal.} Most I2I translations methods are deterministic, since, at inference time,  they can produce only {\em one} translated image $\hat{\pmb{x}}$ given a source image $\pmb{x}$ and a target domain $j$. However, in many practical applications, it is desirable that the appearance of $\hat{\pmb{x}}$ depends also on some random factor, in such a way to be able to produce different plausible translations.\vspace{-0.3\baselineskip}
\end{enumerate}
There are mainly two mechanisms that can be used to obtain 
a specific style code $\pmb{s} \in \pmb{\mathcal{S}}_j$.
The first option is to sample a random vector (e.g., $\pmb{z}\sim\mathcal{N}(\pmb{0},\pmb{I})$) and then use an MLP to transform $\pmb{z}$ into a style code:
 $\pmb{s} = M(\pmb{z}, j)$ \cite{karras2019style}, where   $j$ is the domain label.
The second option is based on extracting the code from  a reference image ($\pmb{x}'\in \pmb{\mathcal{X}}_j$) by means of  an encoder: $\pmb{s} = E(\pmb{x}')$. In our case, we use both of them.

\section{Method}
\label{sec:method}

\Cref{Fig:interpolation-intro} shows the main intuition behind our method. A style space in which different domains are well separated (i.e., disentangled) may not be sufficient to guarantee smooth inter-domain interpolations. When the domain-specific distributions are too far apart from each other, this may lead to what we call ``training gaps'', 
i.e., portions of the space that are not populated with training samples. 
Consequently, at training time, the network has not observed samples in those regions, and, at inference time, it may misbehave when sampling in those regions (e.g., producing image artifacts). Moreover, a non-compact style space may create intra-domain ``training gaps'', leading to the generation of non-realistic images when drawing style codes in these areas.
Thus, we argue that smoothness is related to reducing these training gaps and compacting the latent space.  

Note that the commonly adopted  domain loss~\cite{choi2018stargan} or the multitask adversarial discriminators~\cite{choi2019stargan,liu2019few} might result in domain distributions far apart from each other to facilitate the discriminative task. In order to reduce these training gaps, the domain distributions are expected to be pulled closer while keeping the disentanglement. To achieve these goals, we propose two training losses, described below. 
First, we use a triplet loss \cite{schroff2015facenet} to guarantee the separability of the style codes in different domains. The advantage of the triplet loss is that, using a small margin, the disentanglement of different domains in the latent space can be preserved. Meanwhile, it is convenient to control the inter-domain distance by adjusting the margin.  
However, our empirical results show that the triplet loss alone is insufficient to reduce the training gaps.
For this reason,  we propose to compact style space using a second loss.

We propose to use the Kullback-Leibler (KL) divergence with respect to an a priori Gaussian distribution to make the style space compact. 
This choice is inspired by the regularization adopted in Variational AutoEncoders (VAEs)~\cite{kingma2013auto}. In VAEs, an encoder network is trained to estimate the parameters of a multivariate Gaussian given a single (real) input example. However, in our case, a style code $\pmb{s}$ can be either real (using the encoder $E$, see Sec.~\ref{sec:problem-formulation}) or randomly sampled (using $M$, Sec.~\ref{sec:problem-formulation}), and training an additional encoder to estimate the distribution parameters may be hard and not necessary. For this reason, we propose to simplify the KL divergence using a sample-based $\ell_2$ regularization.

Finally,
as mentioned in Sec.~\ref{sec:problem-formulation}, 
another important aspect of the MMUIT task is content preservation.
To this aim, we propose to use a third loss, based on the idea that the content of an image should be domain-independent  (see  Sec.~\ref{sec:problem-formulation}) and that the similarity of two images with respect to  the content can be estimated using a ``perceptual distance''. 
The latter is computed using a  network pre-trained to simulate the human perceptual similarity~\cite{zhang2018unreasonable}.

In Sec.~\ref{sec:Smoothing} we provide the details of these three losses. Note that our proposed losses can be applied to different I2I translation architectures which have an explicit style space (e.g., a style encoder $E$, see Sec.~\ref{sec:problem-formulation}), possibly jointly with other losses.
In Sec.~\ref{sec:Details} we show a specific implementation case, which we used in our experiments and which is inspired to StarGAN v2~\cite{choi2019stargan}. In the Appendix
we show another implementation case based on TUNIT~\cite{baek2020tunit}.

\subsection{Modeling the Style Space}
\label{sec:Smoothing}

\noindent \textbf{Smoothing and disentangling the style space.}
We propose to use a triplet  loss, which is largely used in metric learning \cite{schroff2015facenet,DBLP:conf/nips/Sohn16,Hermans2017InDO,cao2018vggface2}, to preserve the domain disentanglement: 
\begin{equation}
\label{eq.triplet}
    \mathcal{L}_{tri} = \mathbb{E}_{(\pmb{s}_a,\pmb{s}_p,\pmb{s}_n)\sim\pmb{\mathcal{S}}}[\max\left(||\pmb{s}_a - \pmb{s}_p)|| - ||\pmb{s}_a - \pmb{s}_n|| + \alpha, 0\right)],
\end{equation}
where $\alpha$ is a constant margin and $\pmb{s}_a$ and $\pmb{s}_p$ (i. e., the {\em anchor} and the {\em positive}, adopting the common terminology of the triplet loss \cite{schroff2015facenet}) are style codes extracted from the same domain (e.g., $\pmb{s}_a, \pmb{s}_p \in \pmb{\mathcal{S}}_i$), while the {\em negative} $\pmb{s}_n$ is extracted from a different domain
($\pmb{s}_n \in \pmb{\mathcal{S}}_j, j \neq i$).
These style codes are obtained by sampling real images and
using the encoder. In more detail,
we randomly pick two images from the same domain $i$ ($\pmb{x}_a, \pmb{x}_p \in \pmb{\mathcal{X}}_i$), a third image from another, randomly chosen, domain $j$ ($\pmb{x}_n \in \pmb{\mathcal{X}}_j, j \neq i$), and  then we get the style codes using 
$\pmb{s}_k = E(\pmb{x}_k), k \in \{a, p, n\}$.
Using Eq.~\eqref{eq.triplet}, the network learns to cluster style codes of the same domain. Meanwhile, when the style space is compact, the margin $\alpha$ can control and preserve the disentanglement among the resulting clusters.

Thus, we encourage a compact space forcing an a prior Gaussian distribution on the set of all the style codes $\pmb{\mathcal{S}}$:
\begin{equation}
\label{eq.D-KL}
    \mathcal{L}_{kl} = \mathbb{E}_{\pmb{s}\sim\pmb{\mathcal{S}}}[\mathcal{D}_{\texttt{KL}}(p(\pmb{s})\|\mathcal{N}(\pmb{0}, \pmb{I}))],
\end{equation}
where $\pmb{I}$ is the identity matrix, $\mathcal{D}_{\texttt{KL}}(p\|q)$ is the Kullback-Leibler (KL) divergence and $p(\pmb{s})$ is the distribution corresponding to  the style code  $\pmb{s}$.
However, $p(\pmb{s})$ is unknown. In VAEs, $p(\pmb{s})$ is commonly estimated assuming a Gaussian shape and using an encoder to regress the mean and the covariance-matrix parameters of each single sample-based distribution ~\cite{kingma2013auto}.
Very recently, Ghosh et al. \cite{DBLP:conf/iclr/GhoshSVBS20} showed that, assuming the variance to be constant for all the samples,
the  KL divergence regularization can be simplified (up to a constant) to $\mathcal{L}_{SR}^{CV} (\pmb{x}) = ||\bm{\mu}(\pmb{x})||_2^2$, where ``CV'' stands for Constant-Variance, and $\bm{\mu}(\pmb{x})$ is the mean estimated by the encoder using  $\pmb{x}$. In this paper we propose a  further simplification based on the assumption that 
$\bm{\mu}(\pmb{s}) = \pmb{s}$ (which is reasonable if $\bm{\mu}$ is estimated using only one sample) and we eventually get the proposed 
{\em Style Regularization} (SR) loss: 
\begin{equation}
\label{eq.D-KL-final}
    \mathcal{L}_{SR} = \mathbb{E}_{\pmb{s}\sim\pmb{\mathcal{S}}}[|| \pmb{s} ||_2^2].
\end{equation}
Eq.~\eqref{eq.D-KL-final} penalizes samples $\pmb{s}$ with a large $\ell_2$ norm, so  encouraging the distribution of $\pmb{\mathcal{S}}$ to be a shrunk Gaussian centered on the origin. 
Intuitively, while the SR loss compacts the space, 
 the triplet loss avoids a domain entanglement in the compacted region (see also the Appendix).
Finally, we describe below how the style-code samples are drawn in Eq.~\eqref{eq.D-KL-final}  ($\pmb{s}\sim\pmb{\mathcal{S}}$).
We use a mixed strategy, including both real  and randomly generated codes. More in detail, with probability 0.5, we use a real sample 
$\pmb{x} \in \pmb{\mathcal{X}}$ and we get: $\pmb{s} = E(\pmb{x})$, and, with probability 0.5, we use 
$\pmb{z}\sim\mathcal{N}(\pmb{0},\pmb{I})$ and $\pmb{s} = M(\pmb{z}, j)$. In practice, we alternate mini-batch iterations in which we use only real samples 
with iterations in which we use only generated samples. 

\noindent \textbf{Preserving the source content.}
The third loss we propose aims at preserving the content in the I2I translation: 
\begin{equation}
\label{eq.lpips}
    \mathcal{L}_{cont} = \mathbb{E}_{\pmb{x}\sim\pmb{\mathcal{X}}, \pmb{s}\sim\pmb{\mathcal{S}}} [\psi( \pmb{x}, G(\pmb{x}, \pmb{s}))],
\end{equation}
where $\psi(\pmb{x}_1, \pmb{x}_2)$ estimates the perceptual distance  between $\pmb{x}_1$ and $\pmb{x}_2$ using an externally pre-trained network. 
The rationale behind Eq.~\eqref{eq.lpips} is that, given a source image $\pmb{x}$ belonging to domain $\pmb{\mathcal{X}}_i$, for each
style code $\pmb{s}$, extracted from the set of   {\em all} the domains $\pmb{\mathcal{S}}$,
  we want to minimize the perceptual distance between $\pmb{x}$ and the transformed image $G(\pmb{x}, \pmb{s})$.
By minimizing Eq.~\eqref{eq.lpips},  the perceptual content (extracted through $\psi(\cdot)$)  is encouraged to be independent of the domain (see the definition of content preservation in Sec.~\ref{sec:problem-formulation}). 
Although different perceptual distances can be used (e.g., the Euclidean distance on VGG features~\cite{johnson2016perceptual}),  we implement $\psi(\pmb{x}_1, \pmb{x}_2)$ using the Learned Perceptual Image Patch Similarity (LPIPS) metric~\cite{zhang2018unreasonable}, which was shown to be 
 well aligned with the human perceptual similarity~\cite{zhang2018unreasonable} and it is obtained using  a multi-layer representation of the two input images ($\pmb{x}_1, \pmb{x}_2$) in a pre-trained network.
 
 The sampling procedure in the {\em content preserving} loss ($\mathcal{L}_{cont}$) is similar to the 
SR loss. First, we randomly sample $\pmb{x} \in \pmb{\mathcal{X}}$. Then, we either sample a different reference image 
$\pmb{x}' \in \pmb{\mathcal{X}}$ and get $\pmb{s} = E(\pmb{x}')$, or we use $\pmb{z}\sim\mathcal{N}(\pmb{0},\pmb{I})$ and $\pmb{s} = M(\pmb{z}, j)$.

We sum together
the three proposed losses  and we get:
\begin{equation}
    \mathcal{L}_{smooth} = \mathcal{L}_{cont} + \lambda_{sr} \mathcal{L}_{SR} +  \mathcal{L}_{tri},
\end{equation}
where $\lambda_{sr}$ is the SR loss-specific weight. 

\subsection{Smoothing the Style Space of an Existing Model}
\label{sec:Details}

The proposed $\mathcal{L}_{smooth}$ can be  plugged in existing MMUIT methods which have an explicit style space, by summing it with their original objective function ($\mathcal{L}_{orig}$):
\begin{equation}
    \mathcal{L}_{new} = \mathcal{L}_{smooth} + \mathcal{L}_{orig}.
\end{equation}

In this subsection, we show an example in which $\mathcal{L}_{orig}$ is the original loss of the MMUIT state-of-the-art StarGAN v2~\cite{choi2019stargan}. In the Appendix we show another example  based on TUNIT~\cite{baek2020tunit}, which is the state of the art of fully-unsupervised image-to-image translation. 

In StarGAN v2, the original loss is:
\begin{equation}
\label{eq.LStarGanV2}
    \mathcal{L}_{orig} =   \lambda_{sty}\mathcal{L}_{sty} - \lambda_{ds}\mathcal{L}_{ds} + \lambda_{cyc}\mathcal{L}_{cyc} + \mathcal{L}_{adv}
\end{equation}
where $\lambda_{sty}, \lambda_{ds}$ and $\lambda_{cyc}$ control the contribution of the \emph{style reconstruction}, the \emph{diversity sensitive}, and the \emph{cycle consistency} loss, respectively.

The \textit{style reconstruction} loss~\cite{huang2018multimodal,zhu2017toward,choi2019stargan} pushes  the target  code ($\pmb{s}$) and the code extracted  from the generated image ($E(G(\pmb{x},\pmb{s}))$)  to be as close as possible:
\begin{equation}
    \mathcal{L}_{sty} = \mathbb{E}_{\pmb{x}\sim \pmb{\mathcal{X}}, \pmb{s}\sim\pmb{\mathcal{S}}} \left[ \|\pmb{s} - E(G(\pmb{x},\pmb{s}))\|_1\right].
\end{equation}
The \textit{diversity sensitive} loss~\cite{choi2019stargan, mao2019mode} encourages $G$ to produce diverse images:
\begin{equation}
    \mathcal{L}_{ds} = \mathbb{E}_{\pmb{x}\sim\pmb{\mathcal{X}}_i,  (\pmb{s}_1, \pmb{s}_2)\sim\pmb{\mathcal{S}}_j} \left[\| G(\pmb{x}, \pmb{s}_1) - G(\pmb{x}, \pmb{s}_2 ) \|_1\right].
\end{equation}
The \textit{cycle consistency}~\cite{zhu2017unpaired,choi2018stargan,choi2019stargan} loss is used 
to preserve the  content of the source image $\pmb{x}$:
\begin{equation}
\label{eq:cyc_pixelwise}
    \mathcal{L}_{cyc} = \mathbb{E}_{\pmb{x}\sim \pmb{\mathcal{X}}, \pmb{s}\sim\pmb{\mathcal{S}}}\left[\|\pmb{x} - G(G(\pmb{x}, \pmb{s}), E(\pmb{x}))\|_1\right].
\end{equation}

Finally, StarGAN v2 uses a multitask discriminator~\cite{liu2019few} $D$, which consists of multiple output branches.  Each branch $D_j$ learns a binary classification determining whether an image $\pmb{x}$ is a real image of its dedicated domain $j$ or a fake image.
Thus, the {\em adversarial} loss can be formulated as:
\begin{equation}
    \mathcal{L}_{adv} = \mathbb{E}_{\pmb{x}\sim\pmb{\mathcal{X}}_i,  \pmb{s}\sim\pmb{\mathcal{S}}_j} [\log D_i(\pmb{x}) + \log(1-D_j(G(\pmb{x}, \pmb{s})))]
\end{equation}
Note that this loss encourages the separation of the domain-specific distributions without controlling the relative inter-domain distance (Sec.~\ref{sec:method}). We use it jointly with our $\mathcal{L}_{tri}$.

We refer the reader to \cite{choi2019stargan} and to the Appendix for additional details. In Sec.~\ref{sec:experiments}
we evaluate the combination of our $\mathcal{L}_{smooth}$ with StarGAN v2 (Eq.~\eqref{eq.LStarGanV2}), while in the Appendix we show additional experiments in which 
$\mathcal{L}_{smooth}$ is combined with TUNIT~\cite{baek2020tunit}.

\section{Evaluation Protocols}
\label{sec:metrics}

\noindent\textbf{FID.} For each translation $\pmb{\mathcal{X}}_i \rightarrow \pmb{\mathcal{X}}_j$, we use 1,000 test images and estimate the Fr\'echet Inception Distance (FID)~\cite{NIPS2017_7240} using interpolation results. In more detail, for each image,
we randomly sample  two  style codes ($\pmb{s}_1\in\pmb{\mathcal{S}}_i$ and $\pmb{s}_2\in\pmb{\mathcal{S}}_j$), which are linearly interpolated using 20
points. Each point (included  $\pmb{s}_1$ and $\pmb{s}_2$) is used to generate a translated image. The FID values are computed using the $20 \times 1,000$ outputs. 
A lower FID score indicates a lower discrepancy between the image quality of the real and generated images.

\noindent \textbf{LPIPS.} For a given domain $\pmb{\mathcal{X}}_i$,
we use 1,000 test images  $\pmb{x} \in \pmb{\mathcal{X}}_i$, and, for each $\pmb{x}$, we randomly generate 10 image translations in the target domain $\pmb{\mathcal{X}}_j$. Then, the LPIPS~\cite{zhang2018unreasonable} distances among the 10 generated images are computed. Finally, all distances are averaged.
A higher LPIPS distance indicates a greater  diversity among the generated images. Note that the LPIPS distance ($\psi(\pmb{x}_1, \pmb{x}_2)$) is computed using an {\em externally pre-trained} network~\cite{zhang2018unreasonable}, which is the same we use in Eq.~\eqref{eq.lpips} at training time.

\noindent\textbf{FRD.} For the specific case of face translations, we use a metric based on a pretrained VGGFace2 network ($\phi$)~\cite{schroff2015facenet,cao2018vggface2}, which estimates the visual distance between two faces.
Note that the identity of a person may be considered as a specific case of ``content'' (Sec.~\ref{sec:problem-formulation}). 
We call this metric the Face Recognition Distance (FRD): 
\begin{equation}
    \text{FRD} = \mathbb{E}_{\pmb{x} \sim \pmb{\mathcal{X}}, \pmb{s} \sim \pmb{\mathcal{S}}} \left[\|\phi(\pmb{x}) - \phi(G(\pmb{x}, \pmb{s})))\|_2^2\right].
\end{equation}

\noindent\textbf{PS.} Karras et al.~\cite{karras2019style} recently proposed the Perceptual Path Length (PPL) to evaluate the smoothness and the disentanglement of a semantic latent space. PPL is based on measuring the LPIPS distance between close points in the style space. However, one issue with the PPL is that it can be minimized by a collapsed generator. For this reason,  we alternatively propose the Perceptual Smoothness (PS)
metric, which returns a normalized score in $[0, 1]$, indicating the smoothness of the  style space.

In more detail, let $\pmb{s}_0$ and $\pmb{s}_T$ be  two  codes randomly sampled from the style space, $P = ( \pmb{s}_0, \pmb{s}_1, \ldots, \pmb{s}_T )$  the sequence of the
 linearly interpolated points  between $\pmb{s}_0$ and $\pmb{s}_T$,
 and $A= ( G(\pmb{x}, \pmb{s}_0),  \ldots, G(\pmb{x}, \pmb{s}_T) )$ the corresponding sequence of images generated starting from a source image $\pmb{x}$.
 We measure the degree of
 linear \emph{alignment} of the generated images using:
\begin{equation}
    \ell_{\texttt{alig}} = \mathbb{E}_{\pmb{x} \sim \pmb{\mathcal{X}}, \pmb{s}_0, \pmb{s}_T \sim \pmb{\mathcal{S}}}
    \left[ \frac{\delta(\pmb{x}, \pmb{s}_0, \pmb{s}_T)}{\sum_{t=1}^{T}  \delta(\pmb{x}, \pmb{s}_{t-1}, \pmb{s}_{t})} \right]
\end{equation}
where  $\delta(\pmb{x}, \mathbf{s}_1, \mathbf{s}_2) = \psi(G(\pmb{x}, \mathbf{s}_1), G(\pmb{x}, \mathbf{s}_2))$ and $\psi(\cdot, \cdot)$ is the
LPIPS distance (modified to be a proper metric, more details in the Appendix). 
When $\ell_{\texttt{alig}} = 1$, then the perceptual distance between $G(\pmb{x}, \pmb{s}_0)$ and $G(\pmb{x}, \pmb{s}_T)$ is equal to the sum of the  perceptual distances between consecutive elements in $A$, thus, the  images in $A$
lie along a line in the space of $\psi(\cdot, \cdot)$ (which represents 
 the human perceptual similarity~\cite{zhang2018unreasonable}).
Conversely,  when $\ell_{\texttt{alig}} < 1$, then the images in $A$ contain some  visual attribute not contained in any of the endpoints. For example,  transforming  a short-hair male person to a short-hair girl, we may have $\ell_{\texttt{alig}} < 1$ when the images in $A$  contain people with long hair.
However, although aligned, the images in $A$ may have a non-uniform distance, in which $\delta(\pmb{x}, \pmb{s}_{t-1}, \pmb{s}_{t})$ varies depending on $t$. In order to measure the \emph{uniformity} of these distances, we use
the opposite of the Gini inequality coefficient~\cite{gini1912variabilita}:
\begin{dmath*}
    \resizebox{\columnwidth}{!}{$
    \ell_{\texttt{uni}} = \mathbb{E}_{\subalign{\pmb{x} \sim \pmb{\mathcal{X}}\\\pmb{s}_0, \pmb{s}_T \sim \pmb{\mathcal{S}}}}\left[ 1- \frac{\sum_{i,j=1}^{T} |\delta(\pmb{x}, \pmb{s}_{i-1}, \pmb{s}_i) - \delta(\pmb{x}, \pmb{s}_{j-1}, \pmb{s}_j)|}{2 T^2 \mu_P}  \right]
    $}
\end{dmath*}
where $\mu_P$ is the average value of $\delta(\cdot)$ computed over all the pairs of elements in $P = ( \pmb{s}_0,  \ldots, \pmb{s}_T )$.
Intuitively, $\ell_{\texttt{uni}} = 1$ when 
an evenly-spaced linear interpolation of the style codes corresponds to
 constant changes in the perceived difference of the generated images,
  while $\ell_{\texttt{uni}} = 0$ when there is only one abrupt change in a single step.
Finally, we define PS as the harmonic mean of $\ell_{\texttt{alig}}$ and $\ell_{\texttt{uni}}$:
\begin{equation}
    \text{PS} = 2 \cdot \frac{\ell_{\texttt{alig}} \cdot \ell_{\texttt{uni}}}{\ell_{\texttt{alig}} + \ell_{\texttt{uni}}} \in [0,1].
\end{equation}

\section{Experiments}
\label{sec:experiments}

\begin{figure*}[ht]	
	\renewcommand{\tabcolsep}{1pt}
	\renewcommand{\arraystretch}{0.8}
	\centering
	\footnotesize
	\begin{tabular}{ccc}
		\includegraphics[width=0.92\linewidth]{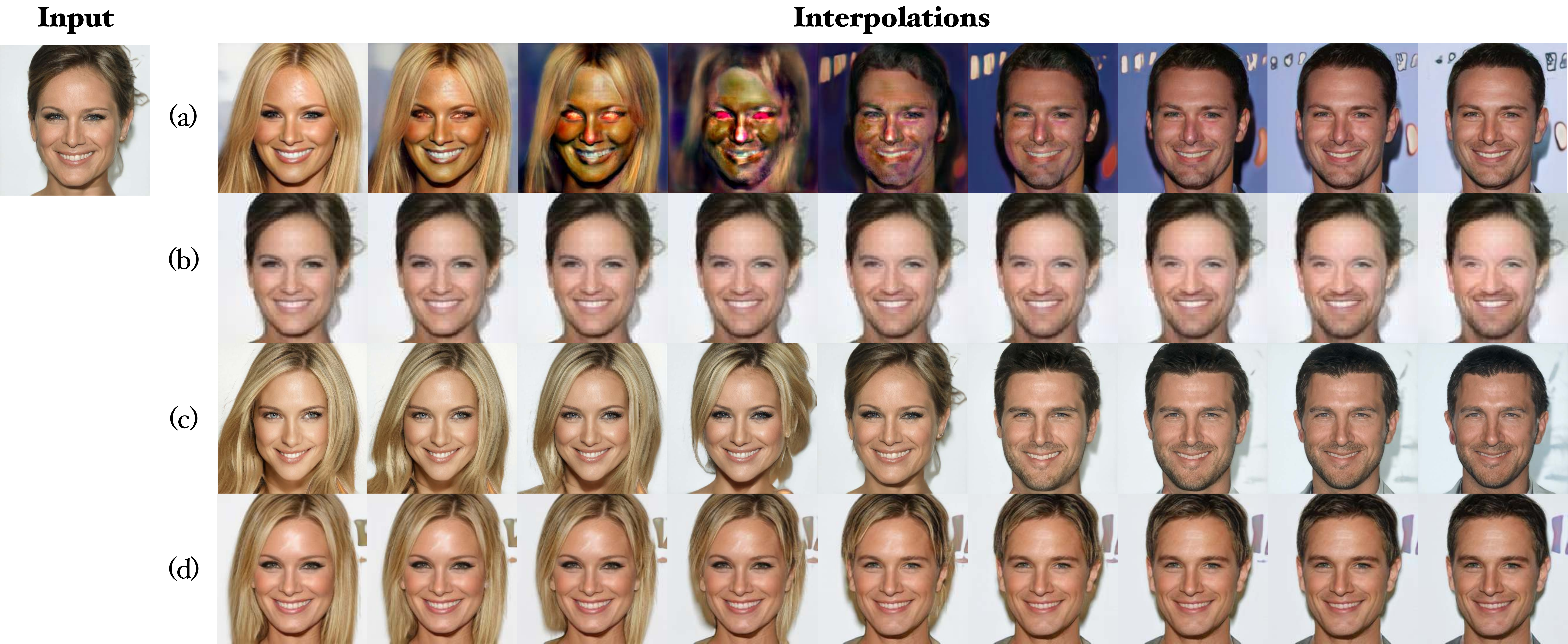}\\
	\end{tabular}
	\vspace{-0.5em}
	\caption{Inter-domain  interpolation results: (a) StarGAN v2~\cite{choi2019stargan}, (b) HomoGAN~\cite{chen2019homomorphic}, (c) InterFaceGAN~\cite{shen2020interpreting}, (d) ours. The domains correspond to genders. Our method generates  smoother results while better preserving the source-person identity. 
	}
\vspace{-4mm}
	\label{Fig:interpolation-celebAHQ}
\end{figure*}

\noindent\textbf{Baselines}. 
We compare our method with three state-of-the-art approaches: (1) StarGAN v2~\cite{choi2019stargan}, the state of the art for the MMUIT task; (2) HomoGAN~\cite{chen2019homomorphic}; and (3) TUNIT~\cite{baek2020tunit}. 
Moreover, as a reference for a high image quality, we also use InterFaceGAN~\cite{shen2020interpreting}, a StyleGAN-based method (trained with $1024 \times 1024$ images) which interpolates the pre-trained semantic space of StyleGAN~\cite{karras2019style} (see Sec.~\ref{sec:related_work}). InterFaceGAN is not designed for domain translation and for preserving the source content, but it can linearly interpolate  a fixed latent space, massively trained with high-resolution images.
All the baselines are tested using the original publicly available codes.

\noindent\textbf{Datasets.} We follow the experimental protocol of StarGAN v2~\cite{choi2019stargan} and we use the CelebA-HQ~\cite{karras2017progressive} and the AFHQ dataset~\cite{choi2019stargan}. The domains are: \emph{male}-\emph{female}, \emph{smile}-\emph{no smile}, \emph{young}-\emph{non young}
in CelebA-HQ; \emph{cat}, \emph{dog}, and \emph{wildlife} in AFHQ. For a fair comparison, all models (except InterFaceGAN) are trained with $256 \times 256$ images. Additional details are provided in the Appendix.

\noindent \textbf{Settings.} We test our method 
in two experimental settings, respectively called ``unsupervised'' (with only set-level annotations) and ``truly unsupervised'' (no annotations~\cite{baek2020tunit}). Correspondingly, we plug our training losses ($\mathcal{L}_{smooth}$) in 
the state-of-the art StarGAN v2~\cite{choi2019stargan} and TUNIT~\cite{baek2020tunit}
(see Sec.~\ref{sec:Smoothing}).
In each setting, we plug our  method in the original architecture without adding additional modules and adopting the original hyper-parameter values without tuning. We refer to the Appendix for more details.

\begin{table}[!ht]
    \renewcommand{\tabcolsep}{2pt}
    \renewcommand{\arraystretch}{1}
    \small
	\centering
	\begin{threeparttable}
	\begin{tabularx}{\columnwidth}{@{}Xrrr rrr@{}}
	\toprule
	\multirow{2}{*}{\textbf{Model}} & \multicolumn{3}{c}{\textbf{PS}$\uparrow$} &  \multicolumn{3}{c}{\textbf{FRD}$\downarrow$}  \\ \cmidrule(lr){2-4} \cmidrule(lr){5-7}
	& Gender & Smile & Age & Gender & Smile & Age \\ \midrule
	HomoGAN~\cite{chen2019homomorphic} & .401 & .351 & .389 & .903 & .820 & .842\\ 
	StarGAN v2~\cite{choi2019stargan} & .272 & .282 & .283  & 1.082 & .894 & .882 \\ 
	Ours & \textbf{.504} & \textbf{.513} & \textbf{.601} & \textbf{.837} & \textbf{.625} & \textbf{.650} \\
	\midrule \midrule
	InterFaceGAN~\cite{shen2020interpreting}\tnote{\S} & .328 & .436 & .409 & .884 & .560 & .722 \\ 
	\bottomrule
	\end{tabularx}
	\end{threeparttable}
	\vspace{-0.5em}
	\caption{Smoothness degree and identity preservation on the CelebA-HQ dataset.  \textsuperscript{\S}Trained on $1024\times1024$ images.
	}\vspace{-2mm}
	\label{tab:celebahq}
\end{table}

\begin{table}[!ht]
    \renewcommand{\tabcolsep}{2pt}
    \renewcommand{\arraystretch}{1}
    \small
	\centering
	\begin{threeparttable}
	\begin{tabularx}{\columnwidth}{@{}Xrrr rrr@{}}
	\toprule
	\multirow{2}{*}{\textbf{Model}} & \multicolumn{3}{c}{\textbf{FID}$\downarrow$} &  \multicolumn{3}{c}{\textbf{LPIPS}$\uparrow$}  \\ \cmidrule(lr){2-4} \cmidrule(lr){5-7}
	& Gender & Smile & Age & Gender & Smile & Age \\ \midrule
	HomoGAN~\cite{chen2019homomorphic} & 55.23 & 58.02 & 57.50 & .010 & .005 & .008  \\ 
	StarGAN v2~\cite{choi2019stargan} & 48.35 & 29.65 & 26.60 & \textbf{.442} & \textbf{.413} & \textbf{.407}  \\
	Ours & \textbf{23.37} & \textbf{22.21} & \textbf{23.57} & .337 & .095 & .128 \\ 
	\midrule \midrule
	InterFaceGAN~\cite{shen2020interpreting}\tnote{\S} & 13.75 & 12.81 & 12.25 & .211 & .115 & .146 \\ 
	\bottomrule
	\end{tabularx}
	\end{threeparttable}
	\vspace{-0.5em}
	\caption{Image quality and translation diversity on the CelebA-HQ dataset.  \textsuperscript{\S}Trained on $1024\times1024$ images.}
	\vspace{-4mm}
	\label{tab:celebahq-quality}
\end{table}

\subsection{Smoothness of the Style Space}
\Cref{Fig:interpolation-celebAHQ} shows a qualitative evaluation using  the style-space interpolation between a source image and a reference style. 
As mentioned in Sec.~\ref{sec:introduction} and \ref{sec:method},
StarGAN v2 frequently generates artifacts  in inter-domain interpolations  (see \Cref{Fig:interpolation-celebAHQ} (a)). 
 HomoGAN  results are very smooth, but they change very little the one from the other, and the model synthetizes lower quality images (\Cref{Fig:interpolation-celebAHQ} (b)).
InterFaceGAN (\Cref{Fig:interpolation-celebAHQ} (c)) was  trained at a higher image resolution with respect to the other models (ours included). However,  compared to our method (\Cref{Fig:interpolation-celebAHQ} (d)), the interpolation results are less smooth, especially in the middle, while the image quality of both methods is very similar.
Moreover, comparing our approach to StarGAN v2,
 our method better preserves the background content in all the generated images.
 
These results are quantitatively confirmed in \Cref{tab:celebahq}. 
The  PS scores show that our proposal improves the state of the art significantly,
which means that it increases the smoothness of the style space in all the CelebA-HQ experiments. 
Note that our results are also better than
InterFaceGAN, whose latent space is  based on the pretrained StyleGAN~\cite{karras2019style}, a very large capacity and training-intensive model.
\Cref{tab:afhq} and \Cref{Fig:interpolation-AFHQ} show similar results also in the challenging  AFHQ dataset, where there is a large inter-domain shift. In this dataset, we tested both the unsupervised   and the truly unsupervised   setting, observing a clear improvement  of both the semantic-space smoothness and the image quality using our method.

The comparison of the qualitative results in \Cref{Fig:interpolation-celebAHQ} and \Cref{Fig:interpolation-AFHQ}
with the PS scores  in \Cref{tab:celebahq} and  \Cref{tab:afhq}, respectively,  show that the proposed PS metric can be reliably used to  evaluate MMUIT models with respect to the style-space smoothness. 
In the Appendix we show additional evidence on the quality  of the PS metrics and how  domain separation can be controlled by tuning the margin value of the triplet loss.

\Cref{tab:celebahq-quality} and \ref{tab:afhq} show that the improvements on the style-space smoothness and the corresponding interpolation results do not come at the expense of the image quality. Conversely, these tables show that the FID values significantly improve with our method.
The LPIPS results in \Cref{tab:celebahq-quality} also show that HomoGAN generates images with little diversity.
However, the LPIPS scores of StarGAN v2 are higher than our method. Nevertheless, the LPIPS metric is influenced by the presence of possible artifacts in the generated images, and, thus, an increased LPIPS value is not necessarily a strength of the model.
We refer to the Appendix for additional qualitative and quantitative results.

Finally, we performed a user study where we
asked 40 users to choose
between the face translations generated by StarGAN v2 and
our method, providing 30 random image pairs to
each user. In $75.8\%$ of cases, the image generated by our
model was selected as the better one, compared to StarGAN
v2 ($25.2\%$).

\subsection{Identity Preservation}

\begin{figure}[t]
\centering
\newcommand{\h}{21.mm}
\newcommand{\hj}{20.mm}
\newcommand{\hdataset}{1mm}
\newcommand{\hreff}{0.5mm}
\newcommand{\himg}{-0.7mm}
\renewcommand{\himg}{-1mm}
\footnotesize{
\makebox[\hj][c]{Source}\hspace{\himg}
\makebox[\hj][c]{Reference}\hspace{\hreff}
\makebox[\hj][c]{StarGAN v2~\cite{choi2019stargan}}\hspace{\himg}
\makebox[\hj][c]{Ours}
}%
\includegraphics[width=\h]{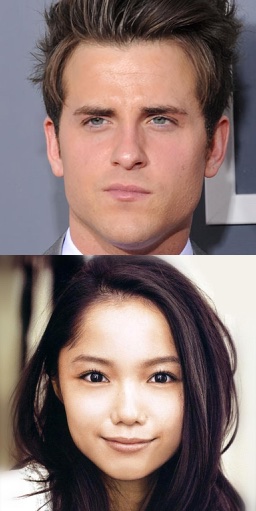}\hspace{\himg}%
\includegraphics[width=\h]{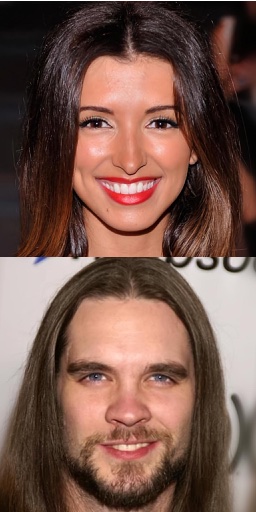}\hspace{\himg}%
\includegraphics[width=\h]{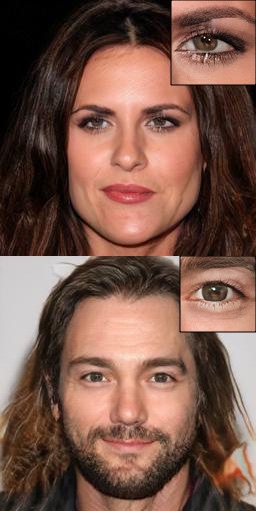}\hspace{\himg}%
\includegraphics[width=\h]{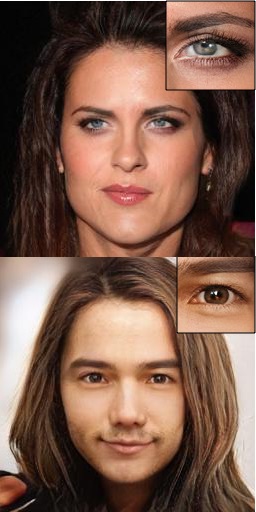}\hspace{\hdataset}%
\vspace{-0.5em}
\caption{Content preservation using  the CelebA-HQ dataset. Our method better preserves the ethnicity and identity of the source images compared to StarGAN v2. 
}
\vspace{-3mm}
\label{fig:refcompare}
\end{figure}

MMUIT models aim at translating images from one  domain to another while keeping the content unchanged.
While this goal is clear, the degree of content preservation is usually evaluated only qualitatively.
Thus, we use the FRD (Sec.~\ref{sec:metrics}) and the most popular I2I translation task (face translation)
to measure the content preservation of the compared models. 
\Cref{tab:celebahq}  shows that our FRD   is the lowest over all the methods compared on the CelebA-HQ dataset, indicating  that our method better maintains the person identity of source images.
Qualitatively, 
\Cref{fig:refcompare} shows that our method  better preserves some distinct face characteristics (e.g., the eye color, the chin shape, or the ethnicity) of the source image while changing the style (i.e., the gender).
This result also suggests that our model might be less influenced by the CelebA-HQ biases (e.g. Caucasian people).
Additional experiments, with similar results, are presented  in the Appendix for smile and age translations.

\subsection{Ablation Study}
In this section, we evaluate the importance of each proposed component. \Cref{tab:ablation} shows the FID,  LPIPS, PS and FRD values for all the configurations, where each component is individually  added to the baseline StarGAN v2, using CelebA-HQ.
First, we observe that adding the $\mathcal{L}_{tri}$ loss to the baseline improves the quality, the diversity and the content preservation of the generated images. However the PS score decreases. This result suggests that better disentanglement might separate too much the styles between domains, thus decreasing the interpolation smoothness.
The addition of $\mathcal{L}_{SR}$ helps improving most of the metrics but the diversity, showing that a more compact style space is a  desirable property for MMUIT.
As mentioned before, we note that higher diversity (LPIPS) might not be strictly related to high-quality images. 

The combination of the two proposed smoothness losses dramatically improves the quality of generated images and the smoothness of the style space. This suggests that the style space should be compact and disentangled, while keeping the  style clusters of different domains close to each other.
Finally,   $\mathcal{L}_{cont}$ further improves the FID, the PS and the FRD scores.
The final configuration corresponds to our full-method and confirms that all the proposed components are helpful.
We refer to the Appendix Sec. B for additional analysis on the  contribution  of  our  losses.

\begin{table}[!ht]
    \renewcommand{\tabcolsep}{3pt}
    \renewcommand{\arraystretch}{1}
    \small
	\centering
	\begin{tabularx}{\columnwidth}{@{}lX rr@{}}
	\toprule
	\textbf{Model} & \textbf{Setting} & \textbf{FID}$\downarrow$ & \textbf{PS}$\uparrow$ \\
	\midrule
	StarGAN v2~\cite{choi2019stargan}& \multirow{2}{*}{Unsupervised} & 15.64 &  .226 \\
	Ours & & \textbf{14.67} &  \textbf{.301} \\ 
	\midrule \midrule
	TUNIT~\cite{baek2020tunit} & \multirow{2}{*}{Truly Unsupervised} & 29.45 &  .443 \\ 
	Ours & & \textbf{16.59} &  \textbf{.447} \\ 
	\bottomrule
	\end{tabularx}
	\vspace{-0.5em}
	\caption{Quantitative evaluation on the AFHQ dataset.
	}\vspace{-3mm}
	\label{tab:afhq}
\end{table} 

\begin{figure}[ht]	
	\renewcommand{\tabcolsep}{1pt}
	\renewcommand{\arraystretch}{0.6}
	\centering
	\footnotesize
	\begin{tabular}{cc}
		(a) & \includegraphics[align=c,width=0.94\columnwidth]{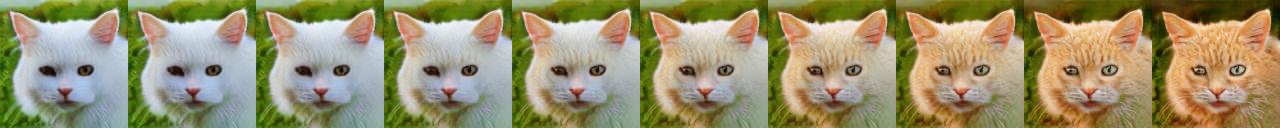}\\%
		(b) & \includegraphics[align=c,width=0.94\columnwidth]{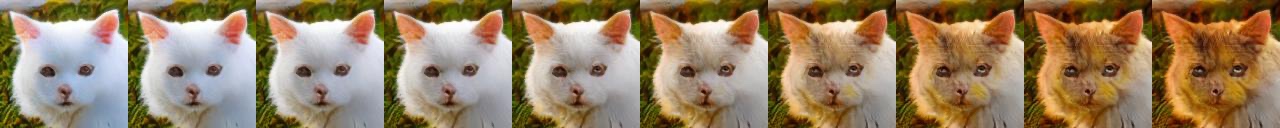}\\%
		(c) & \includegraphics[align=c,width=0.94\columnwidth]{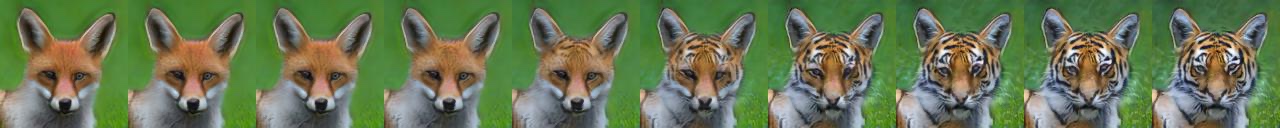} \\
		(d) & \includegraphics[align=c,width=0.94\columnwidth]{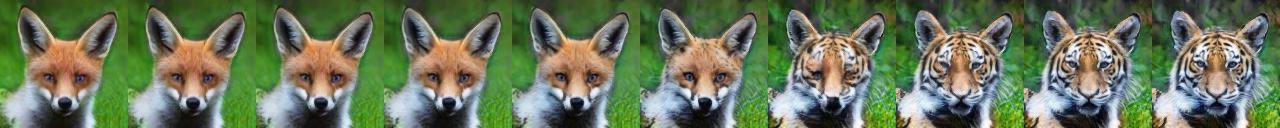}\\%
	\end{tabular}
	\vspace{-0.5em}
	\caption{AFHQ dataset. (b,d) Generation results using  TUNIT~\cite{baek2020tunit}. 
	 (a,c) TUNIT jointly with our losses. 
	} \vspace{-4mm}
	\label{Fig:interpolation-AFHQ}
\end{figure}

\begin{table}[!ht]
    \renewcommand{\tabcolsep}{3pt}
    \renewcommand{\arraystretch}{1}
    \small
	\centering
	\begin{tabularx}{\columnwidth}{@{}Xrrrr@{}}
	\toprule
     \textbf{Model} & \textbf{FID}$\downarrow$ & \textbf{LPIPS}$\uparrow$ & \textbf{PS}$\uparrow$ & \textbf{FRD}$\downarrow$ \\ \midrule
	 A: Baseline StarGAN v2~\cite{choi2019stargan} & 48.35 & \textbf{.442} & .272 & 1.082\\
	 A + $\mathcal{L}_{tri}$ & 37.54 & .403 & .292 & 1.040\\ 
	 A + $\mathcal{L}_{SR}$ & 35.23 & .368 & .432 & .912 \\
	 A + $\mathcal{L}_{SR}$, $\mathcal{L}_{tri}$ & 24.29 & .374 & .501 & .848 \\ 
	 A + $\mathcal{L}_{SR}$, $\mathcal{L}_{tri}$, $\mathcal{L}_{cont}$ & \textbf{23.37} & .337 & \textbf{.504} & \textbf{.837} \\
	\bottomrule
	\end{tabularx}
	\vspace{-0.5em}
	\caption{Ablation study on the CelebA-HQ dataset with a gender translation task.  
	}\vspace{-4mm}
	\label{tab:ablation}
\end{table}

\section{Conclusion}
In this paper, we proposed a new training strategy based on three specific losses which jointly improve both the smoothness of the  style space and the content preservation of existing  MMUIT models. 
We also proposed the PS metric, which specifically evaluates the style smoothness of I2I translation models.
The experimental results show that our method significantly improves  both the smoothness and the quality of the interpolation results and  the  translated images.

\noindent\textbf{Acknowledgements.} This work was supported by EU H2020 SPRING No.871245 and by AI4Media No.951911.

{\small
\bibliographystyle{ieee_fullname}
\bibliography{ref}

\begin{thebibliography}{10}\itemsep=-1pt

\bibitem{alami2018unsupervised}
Youssef Alami~Mejjati, Christian Richardt, James Tompkin, Darren Cosker, and
  Kwang~In Kim.
\newblock Unsupervised attention-guided image-to-image translation.
\newblock {\em NeurIPS}, 2018.

\bibitem{arvanitidis2017latent}
Georgios Arvanitidis, Lars~Kai Hansen, and Søren Hauberg.
\newblock Latent space oddity: on the curvature of deep generative models.
\newblock In {\em ICLR}, 2018.

\bibitem{baek2020tunit}
Kyungjune Baek, Yunjey Choi, Youngjung Uh, Jaejun Yoo, and Hyunjung Shim.
\newblock Rethinking the truly unsupervised image-to-image translation.
\newblock {\em arXiv preprint arXiv:2006.06500}, 2020.

\bibitem{balntas2016learning}
Vassileios Balntas, Edgar Riba, Daniel Ponsa, and Krystian Mikolajczyk.
\newblock Learning local feature descriptors with triplets and shallow
  convolutional neural networks.
\newblock In {\em BMVC}, 2016.

\bibitem{berthelot2018understanding}
David Berthelot, Colin Raffel, Aurko Roy, and Ian Goodfellow.
\newblock Understanding and improving interpolation in autoencoders via an
  adversarial regularizer.
\newblock {\em arXiv preprint arXiv:1807.07543}, 2018.

\bibitem{bojanowski2017optimizing}
Piotr Bojanowski, Armand Joulin, David Lopez-Paz, and Arthur Szlam.
\newblock Optimizing the latent space of generative networks.
\newblock In {\em ICML}, 2018.

\bibitem{cao2018vggface2}
Qiong Cao, Li Shen, Weidi Xie, Omkar~M Parkhi, and Andrew Zisserman.
\newblock Vggface2: A dataset for recognising faces across pose and age.
\newblock In {\em IEEE International Conference on Automatic Face and Gesture
  Recognition (FG)}, 2018.

\bibitem{chen2019homomorphic}
Ying-Cong Chen, Xiaogang Xu, Zhuotao Tian, and Jiaya Jia.
\newblock Homomorphic latent space interpolation for unpaired image-to-image
  translation.
\newblock In {\em CVPR}, 2019.

\bibitem{choi2018stargan}
Yunjey Choi, Minje Choi, Munyoung Kim, Jung-Woo Ha, Sunghun Kim, and Jaegul
  Choo.
\newblock Stargan: Unified generative adversarial networks for multi-domain
  image-to-image translation.
\newblock In {\em CVPR}, 2018.

\bibitem{choi2019stargan}
Yunjey Choi, Youngjung Uh, Jaejun Yoo, and Jung-Woo Ha.
\newblock Stargan v2: Diverse image synthesis for multiple domains.
\newblock In {\em CVPR}, 2020.

\bibitem{DBLP:conf/iclr/GhoshSVBS20}
Partha Ghosh, Mehdi S.~M. Sajjadi, Antonio Vergari, Michael~J. Black, and
  Bernhard Sch{\"{o}}lkopf.
\newblock From variational to deterministic autoencoders.
\newblock In {\em ICLR}, 2020.

\bibitem{gini1912variabilita}
Corrado Gini.
\newblock Variabilit{\`a} e mutabilit{\`a} (variability and mutability).
\newblock {\em Memorie di metodologica statistica}, 1912.

\bibitem{gong2019dlow}
Rui Gong, Wen Li, Yuhua Chen, and Luc~Van Gool.
\newblock Dlow: Domain flow for adaptation and generalization.
\newblock In {\em CVPR}, 2019.

\bibitem{Hermans2017InDO}
Alexander Hermans, Lucas Beyer, and Bastian Leibe.
\newblock In defense of the triplet loss for person re-identification.
\newblock {\em arXiv:1703.07737}, 2017.

\bibitem{NIPS2017_7240}
Martin Heusel, Hubert Ramsauer, Thomas Unterthiner, Bernhard Nessler, and Sepp
  Hochreiter.
\newblock Gans trained by a two time-scale update rule converge to a local nash
  equilibrium.
\newblock In {\em NeurIPS}, 2017.

\bibitem{huang2018multimodal}
Xun Huang, Ming-Yu Liu, Serge Belongie, and Jan Kautz.
\newblock Multimodal unsupervised image-to-image translation.
\newblock In {\em ECCV}, 2018.

\bibitem{isola2017image}
Phillip Isola, Jun-Yan Zhu, Tinghui Zhou, and Alexei~A Efros.
\newblock Image-to-image translation with conditional adversarial networks.
\newblock In {\em CVPR}, 2017.

\bibitem{jahanian2019steerability}
Ali Jahanian, Lucy Chai, and Phillip Isola.
\newblock On the ''steerability" of generative adversarial networks.
\newblock In {\em ICLR}, 2020.

\bibitem{johnson2016perceptual}
Justin Johnson, Alexandre Alahi, and Li Fei-Fei.
\newblock Perceptual losses for real-time style transfer and super-resolution.
\newblock In {\em ECCV}, 2016.

\bibitem{karras2017progressive}
Tero Karras, Timo Aila, Samuli Laine, and Jaakko Lehtinen.
\newblock Progressive growing of gans for improved quality, stability, and
  variation.
\newblock In {\em ICLR}, 2018.

\bibitem{karras2019style}
Tero Karras, Samuli Laine, and Timo Aila.
\newblock A style-based generator architecture for generative adversarial
  networks.
\newblock In {\em CVPR}, 2019.

\bibitem{karras2020analyzing}
Tero Karras, Samuli Laine, Miika Aittala, Janne Hellsten, Jaakko Lehtinen, and
  Timo Aila.
\newblock Analyzing and improving the image quality of stylegan.
\newblock In {\em CVPR}, 2020.

\bibitem{kim2017learning}
Taeksoo Kim, Moonsu Cha, Hyunsoo Kim, Jung~Kwon Lee, and Jiwon Kim.
\newblock Learning to discover cross-domain relations with generative
  adversarial networks.
\newblock In {\em ICML}, 2017.

\bibitem{kingma2013auto}
Diederik~P Kingma and Max Welling.
\newblock Auto-encoding variational bayes.
\newblock {\em arXiv preprint arXiv:1312.6114}, 2013.

\bibitem{krizhevsky2017imagenet}
Alex Krizhevsky, Ilya Sutskever, and Geoffrey~E Hinton.
\newblock Imagenet classification with deep convolutional neural networks.
\newblock {\em Communications of the ACM}, 60(6):84--90, 2017.

\bibitem{gradient1998yann}
Yann Lecun, L\'eon Bottou, Yoshua Bengio, and Patrick Haffner.
\newblock Gradient-based learning applied to document recognition.
\newblock {\em Proceedings of the IEEE}, 86(11):2278--2324, 1998.

\bibitem{lecun1998gradient}
Yann LeCun, L{\'e}on Bottou, Yoshua Bengio, Patrick Haffner, et~al.
\newblock Gradient-based learning applied to document recognition.
\newblock {\em Proceedings of the IEEE}, 86(11):2278--2324, 1998.

\bibitem{lee2018diverse}
Hsin-Ying Lee, Hung-Yu Tseng, Jia-Bin Huang, Maneesh Singh, and Ming-Hsuan
  Yang.
\newblock Diverse image-to-image translation via disentangled representations.
\newblock In {\em ECCV}, 2018.

\bibitem{lee2019drit++}
Hsin-Ying Lee, Hung-Yu Tseng, Qi Mao, Jia-Bin Huang, Yu-Ding Lu, Maneesh Singh,
  and Ming-Hsuan Yang.
\newblock Drit++: Diverse image-to-image translation via disentangled
  representations.
\newblock {\em International Journal of Computer Vision}, 128(10):2402--2417,
  2020.

\bibitem{liu2017unsupervised}
Ming-Yu Liu, Thomas Breuel, and Jan Kautz.
\newblock Unsupervised image-to-image translation networks.
\newblock In {\em NeurIPS}, 2017.

\bibitem{liu2019few}
Ming-Yu Liu, Xun Huang, Arun Mallya, Tero Karras, Timo Aila, Jaakko Lehtinen,
  and Jan Kautz.
\newblock Few-shot unsupervised image-to-image translation.
\newblock In {\em ICCV}, 2019.

\bibitem{10.1145/3394171.3413505}
Yahui Liu, Marco De~Nadai, Deng Cai, Huayang Li, Xavier Alameda-Pineda, Nicu
  Sebe, and Bruno Lepri.
\newblock Describe what to change: A text-guided unsupervised image-to-image
  translation approach.
\newblock In {\em ACM MM}, 2020.

\bibitem{liu2015deep}
Ziwei Liu, Ping Luo, Xiaogang Wang, and Xiaoou Tang.
\newblock Deep learning face attributes in the wild.
\newblock In {\em ICCV}, 2015.

\bibitem{MSGAN}
Qi Mao, Hsin-Ying Lee, Hung-Yu Tseng, Siwei Ma, and Ming-Hsuan Yang.
\newblock Mode seeking generative adversarial networks for diverse image
  synthesis.
\newblock In {\em CVPR}, 2019.

\bibitem{mao2019mode}
Qi Mao, Hsin-Ying Lee, Hung-Yu Tseng, Siwei Ma, and Ming-Hsuan Yang.
\newblock Mode seeking generative adversarial networks for diverse image
  synthesis.
\newblock In {\em CVPR}, 2019.

\bibitem{mo2018instanceaware}
Sangwoo Mo, Minsu Cho, and Jinwoo Shin.
\newblock Instance-aware image-to-image translation.
\newblock In {\em ICLR}, 2019.

\bibitem{richardson2020encoding}
Elad Richardson, Yuval Alaluf, Or Patashnik, Yotam Nitzan, Yaniv Azar, Stav
  Shapiro, and Daniel Cohen-Or.
\newblock Encoding in style: a stylegan encoder for image-to-image translation.
\newblock {\em arXiv preprint arXiv:2008.00951}, 2020.

\bibitem{romero2019smit}
Andr{\'e}s Romero, Pablo Arbel{\'a}ez, Luc Van~Gool, and Radu Timofte.
\newblock Smit: Stochastic multi-label image-to-image translation.
\newblock In {\em ICCV Workshops}, 2019.

\bibitem{schroff2015facenet}
Florian Schroff, Dmitry Kalenichenko, and James Philbin.
\newblock Facenet: A unified embedding for face recognition and clustering.
\newblock In {\em CVPR}, 2015.

\bibitem{shen2020interpreting}
Yujun Shen, Jinjin Gu, Xiaoou Tang, and Bolei Zhou.
\newblock Interpreting the latent space of gans for semantic face editing.
\newblock In {\em CVPR}, 2020.

\bibitem{PAMIStephane}
Aliaksandr Siarohin, St{\'{e}}phane Lathuili{\`{e}}re, Enver Sangineto, and
  Nicu Sebe.
\newblock {Appearance and Pose-Conditioned Human Image Generation using
  Deformable GANs}.
\newblock {\em IEEE TPAMI}, 2020.

\bibitem{DBLP:conf/nips/Sohn16}
Kihyuk Sohn.
\newblock Improved deep metric learning with multi-class n-pair loss objective.
\newblock In {\em NIPS}, 2016.

\bibitem{taigmanPW17}
Yaniv Taigman, Adam Polyak, and Lior Wolf.
\newblock Unsupervised cross-domain image generation.
\newblock In {\em ICLR}, 2017.

\bibitem{wu2019relgan}
Po-Wei Wu, Yu-Jing Lin, Che-Han Chang, Edward~Y Chang, and Shih-Wei Liao.
\newblock Relgan: Multi-domain image-to-image translation via relative
  attributes.
\newblock In {\em ICCV}, 2019.

\bibitem{zhang2018unreasonable}
Richard Zhang, Phillip Isola, Alexei~A Efros, Eli Shechtman, and Oliver Wang.
\newblock The unreasonable effectiveness of deep features as a perceptual
  metric.
\newblock In {\em CVPR}, 2018.

\bibitem{zhu2020domain}
Jiapeng Zhu, Yujun Shen, Deli Zhao, and Bolei Zhou.
\newblock In-domain gan inversion for real image editing.
\newblock In {\em ECCV}, 2020.

\bibitem{zhu2017unpaired}
Jun-Yan Zhu, Taesung Park, Phillip Isola, and Alexei~A Efros.
\newblock Unpaired image-to-image translation using cycle-consistent
  adversarial networks.
\newblock In {\em ICCV}, 2017.

\bibitem{zhu2017toward}
Jun-Yan Zhu, Richard Zhang, Deepak Pathak, Trevor Darrell, Alexei~A Efros,
  Oliver Wang, and Eli Shechtman.
\newblock Toward multimodal image-to-image translation.
\newblock In {\em NeurIPS}, 2017.

\end{thebibliography}
}

\appendix

\section{Model Architecture}

\begin{figure*}[!ht]	
	\centering
	\includegraphics[width=0.92\linewidth]{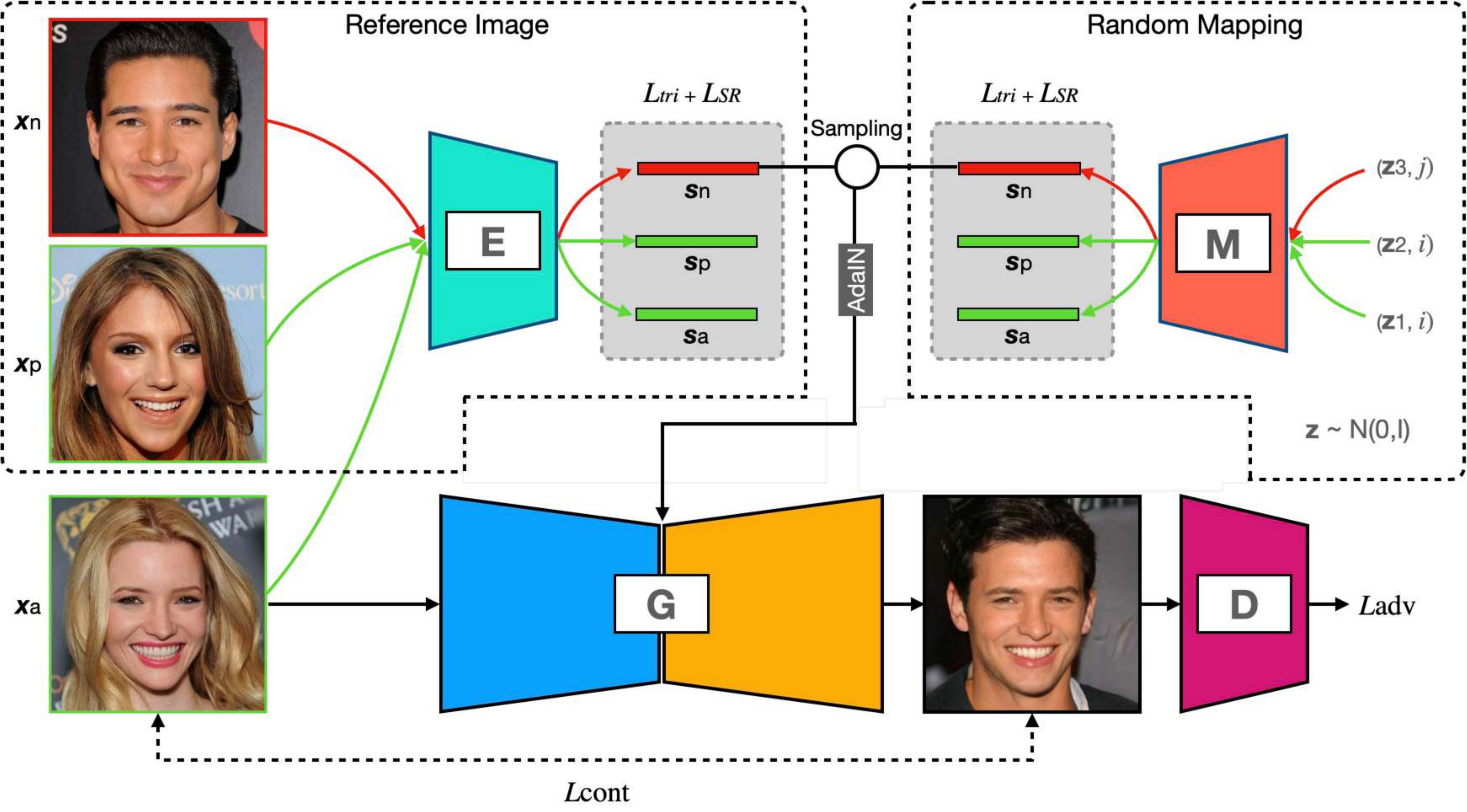}
	\vspace{-1em}
	\caption{Our MMUIT generative  framework and the style-code sampling strategies.}
	\label{Fig:appendix_framework}
\end{figure*}

Fig.~\ref{Fig:appendix_framework} shows the framework of our proposed method for MMUIT tasks. The model is composed of an image generator $G$,
a discriminator $D$, an encoder $E$ and an MLP $M$.
$G$ generates a new image from a source image $\pmb{x}_a$ and a style code $\pmb{s}$, which can either be extracted from a reference image (i.e. $\pmb{s}_p = E(\pmb{x}_p$), or from a randomly sampled vector $\pmb{z}\sim N(0,1)$ through $\pmb{s}_p = M(\pmb{z}$).
The discriminator $D$ learns to classify  an image as either a real image in its associated domain, or a fake image.

As explained in the main paper, we use  $\mathcal{L}_{tri}$, $\mathcal{L}_{SR}$ and $\mathcal{L}_{cont}$  to compact and disentangle the style space
and to help preserving the source content. In Fig.~\ref{Fig:appendix_framework}, 
  $\pmb{s}_n$ is  a style code of a domain different from the domain shared by 
$\pmb{s}_p$ and $\pmb{s}_a$.

\section{Analysing the Style-Space Compactness}

\subsection{Inter-domain Distance Distributions}

\begin{figure*}[ht]	
	\renewcommand{\tabcolsep}{1pt}
	\renewcommand{\arraystretch}{0.8}
	\centering
	\footnotesize
	\begin{tabular}{ccc}
		\includegraphics[width=0.32\linewidth]{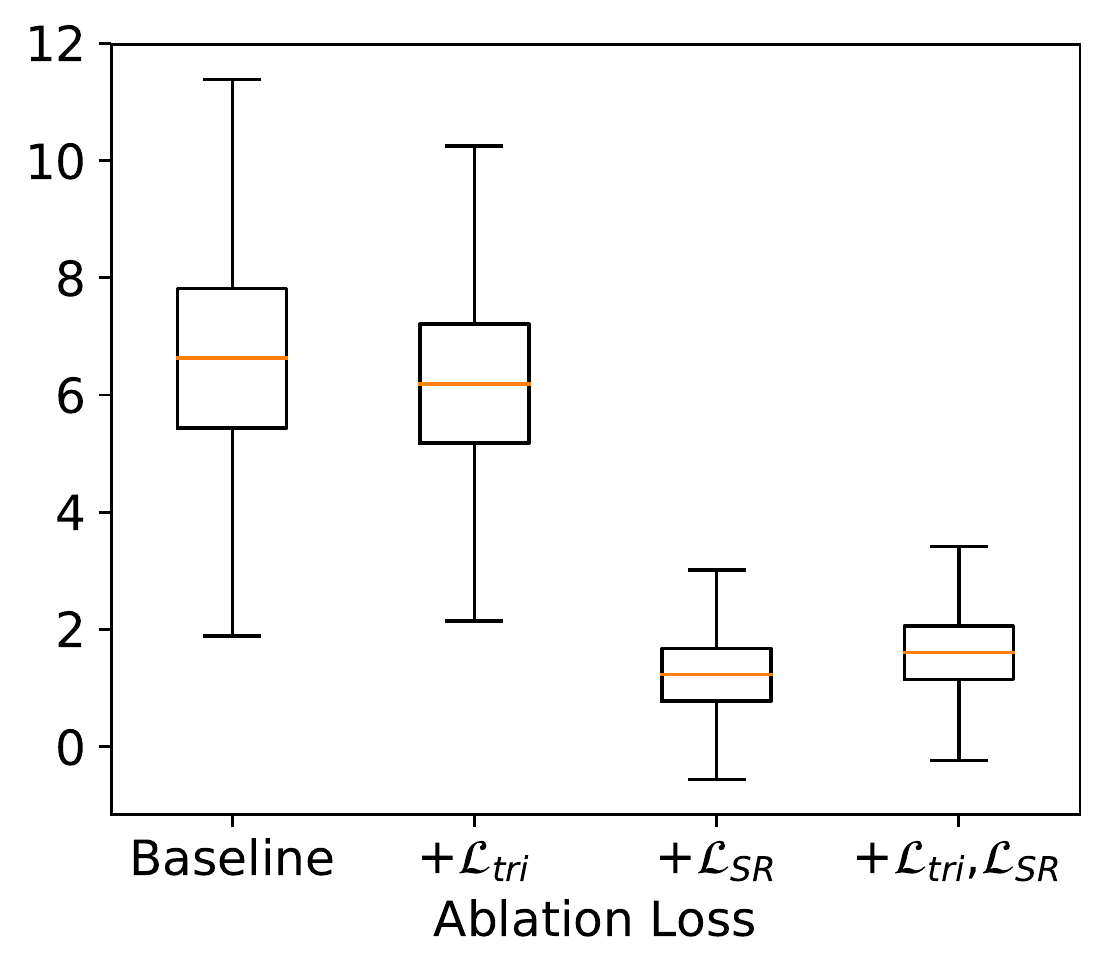} & 
		\includegraphics[width=0.32\linewidth]{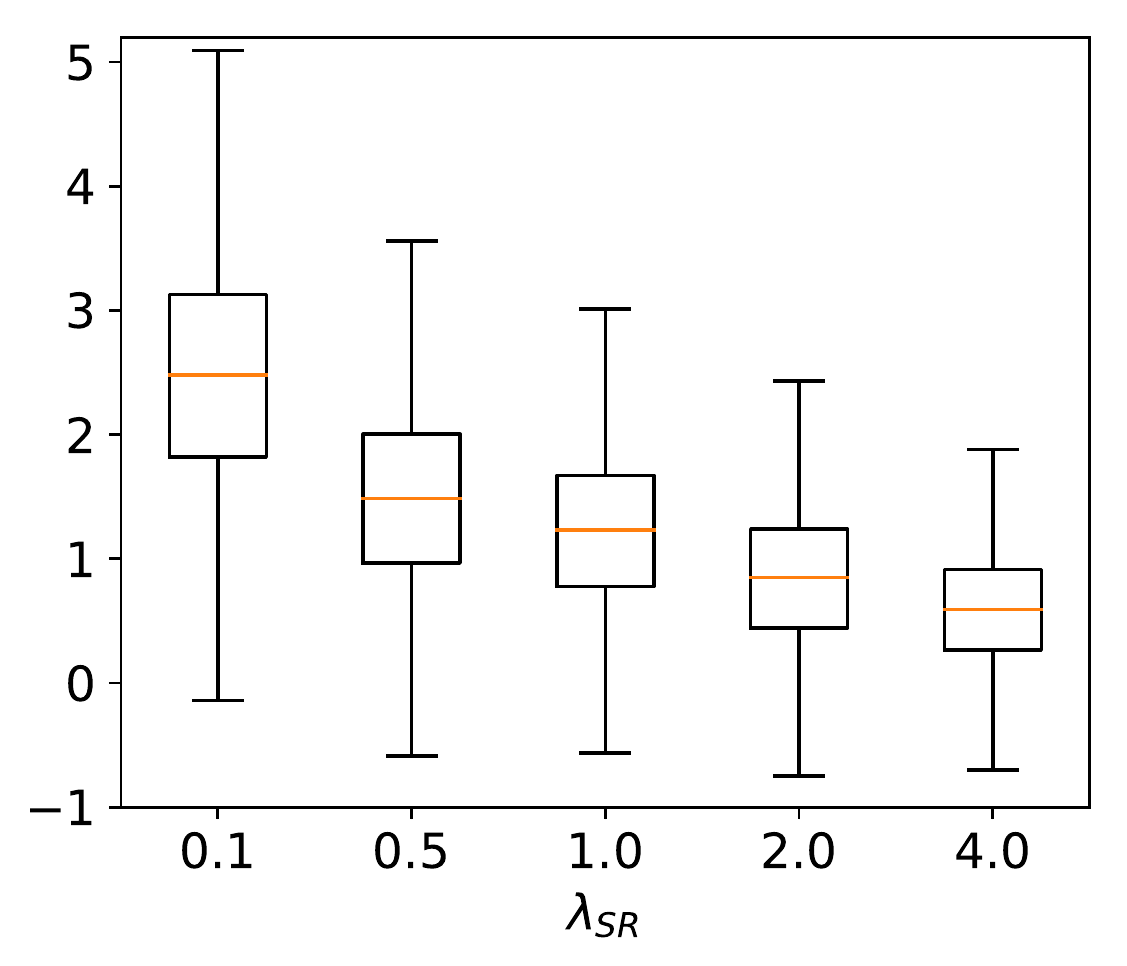} &
		\includegraphics[width=0.32\linewidth]{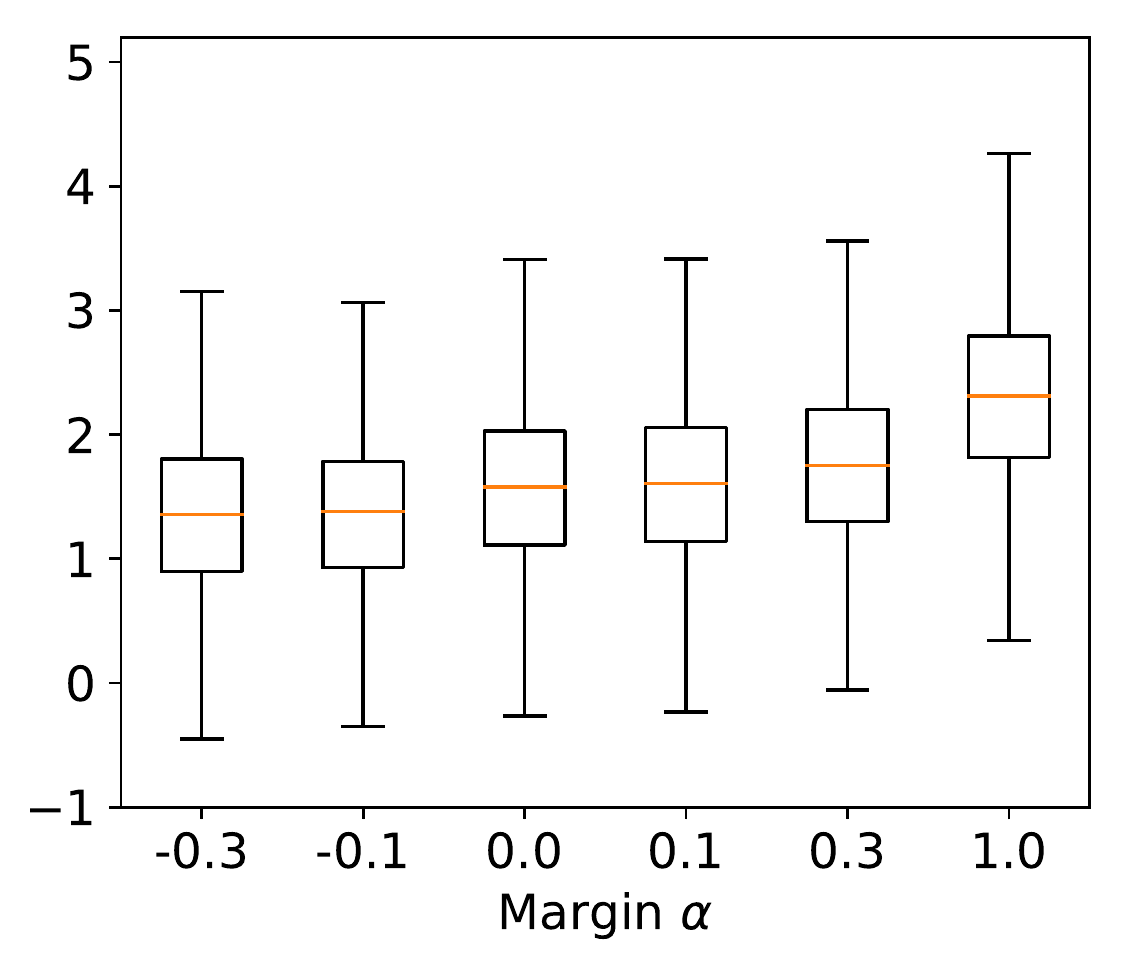} \\
		(a)  & (b) & (c) \\ 
	\end{tabular}
	\vspace{-1em}
	\caption{Distribution of ($d_s(\mathbf{s}_a, \mathbf{s}_n) - d_s(\mathbf{s}_a, \mathbf{s}_p)$) on different experimental settings on the CelebA-HQ dataset.
	(a) shows that  $\mathcal{L}_{SR}$ helps to  compact the style space, while $\mathcal{L}_{tri}$ can adjust the distance between the  style clusters. (b) shows that the weight of the $\mathcal{L}_{SR}$ can control the compactness of the  style space.
(c) shows that increasing the margin $\alpha$ in $\mathcal{L}_{tri}$ has an  effect on the distances between clusters. 
}
	\label{Fig:distance_distribution}
\end{figure*}

In order to estimate the inter-domain distances and 
  the  degree of compactness of a high-dimensional semantic space, we compute the  distribution of
  the distances
  $(d_s(\pmb{s}_a, \pmb{s}_n) - d_s(\pmb{s}_a, \pmb{s}_p))$. Specifically, we use the CelebA-HQ dataset~\cite{karras2017progressive} and we randomly sample 10,000 triplets $(\pmb{s}_a, \pmb{s}_p, \pmb{s}_n)$ where $\pmb{s}_a \sim \pmb{\mathcal{S}}_i$, $\pmb{s}_p \sim \pmb{\mathcal{S}}_i$ and $\pmb{s}_n \sim \pmb{\mathcal{S}}_j$ with $i \neq j$. 
Fig.~\ref{Fig:distance_distribution} shows the distribution of $(d_s(\pmb{s}_a, \pmb{s}_n) - d_s(\pmb{s}_a, \pmb{s}_p))$ under different experimental settings.

Fig.~\ref{Fig:distance_distribution} (a) shows that the distance distribution of the baseline system (without using $\mathcal{L}_{tri}$ and $\mathcal{L}_{SR}$) is relatively wide and corresponds to the largest median. 
Our $\mathcal{L}_{tri}$ loss with a small margin can slightly reduce both the range between the lower quartile to upper quartile and the range between the minimum  to the maximum score. Conversely,  $\mathcal{L}_{SR}$ ($\lambda_{SR} = 1.0$) compacts the space  significantly. Jointly using $\mathcal{L}_{SR}$ and $\mathcal{L}_{tri}$ ($\alpha=0.1$), the    $\mathcal{L}_{SR}$-only distribution is slightly shifted up. 
Fig.~\ref{Fig:distance_distribution} (b) shows the impact of $\lambda_{SR}$ when we use $\mathcal{L}_{SR}$ without $\mathcal{L}_{tri}$. 
Conversely, Fig.~\ref{Fig:distance_distribution} (c) analyses the case of  jointly using $\mathcal{L}_{SR}$ (with $\lambda_{SR} = 1.0$) and $\mathcal{L}_{tri}$
while changing the margin $\alpha$. The latter experiment shows that the Triplet Margin loss can adjust the distance between style clusters, since the ranges between the minimum and the maximum score are shifted when  using a larger  $\alpha$.

\begin{figure}[ht]	
	\centering
		\includegraphics[width=0.96\linewidth]{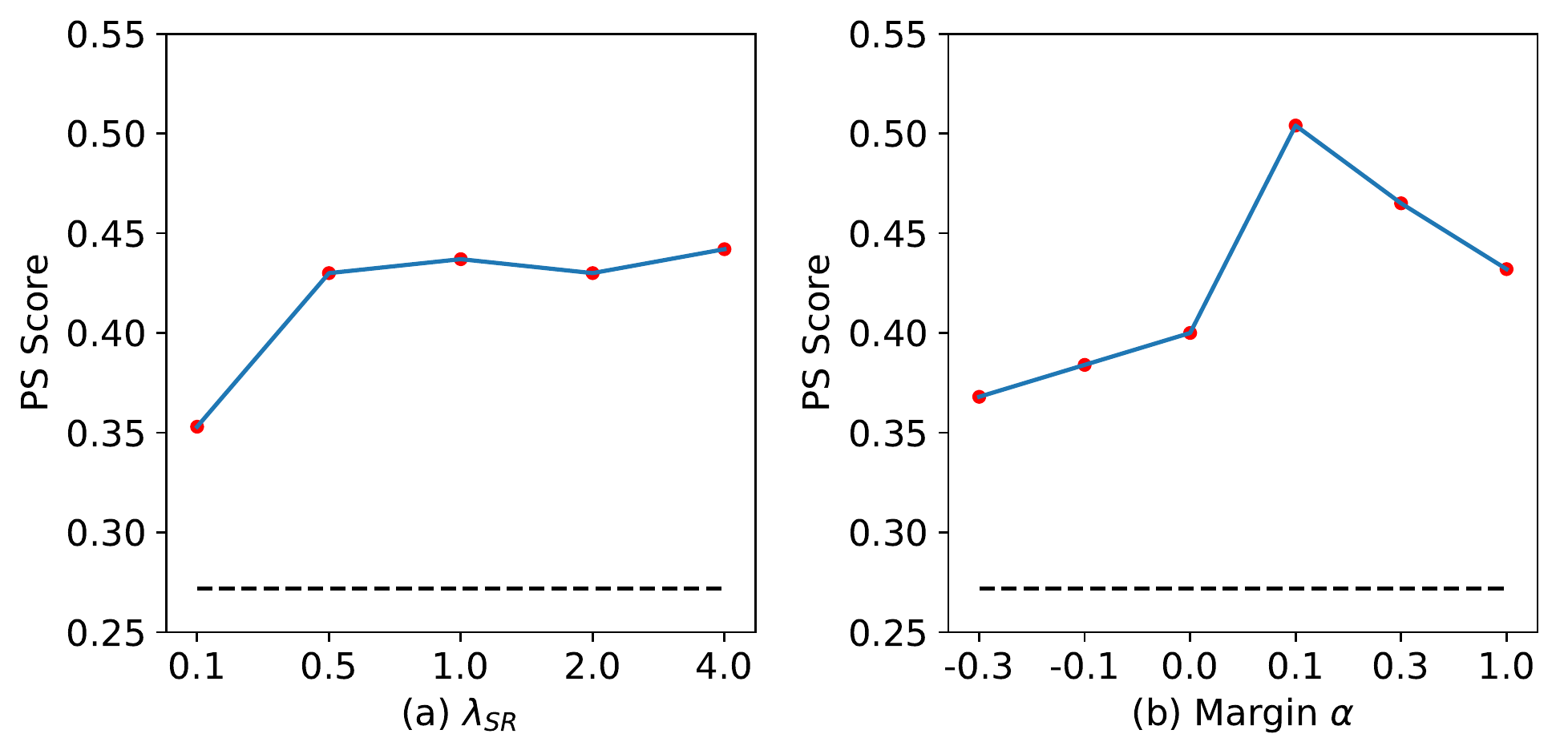}
	\vspace{-0.6em}
	\caption{An ablation study on the influence of both (a) the SR loss weigh $\lambda_{SR}$ and (b) the triplet loss margin $\alpha$ ($\lambda_{SR}=1.0$) in the PS scores. The black dashed line refers to StarGAN v2~\cite{choi2019stargan}.
	}
	\label{Fig:ps_score}
\end{figure}

The corresponding PS scores are presented in Fig.~\ref{Fig:ps_score}, which shows that increasing $\lambda_{SR}$ helps smoothing the space, but when $\lambda_{SR} > 0.5$, only limited improvements are obtained (see Fig.~\ref{Fig:ps_score} (a)). 

As shown in the main paper, the Triplet loss significantly influences the image quality and smoothness of I2I translations. Interestingly, the margin $\alpha$ also plays an important role. Using a small positive margin (e.g., 0.1) is enough to keep the disentanglement and achieve the best PS score, as shown in Fig.~\ref{Fig:ps_score} (b). Meanwhile, a large margin can push the style clusters far away from each other, which may be harmful for the smoothness degree of the space.

\subsection{An Alternative Style Regularization}

A possible alternative to the style-regularization loss ($\mathcal{L}_{SR}$), 
is based on the following formulation, whose goal is to compact the style codes 
close to the surface of the zero-centered, $n$-dimensional unit sphere: 
\begin{equation}
    \mathcal{L}_{sph} = \mathbb{E}_{\pmb{s}\sim\pmb{\mathcal{S}}}\left[|\|\pmb{s}\|_2 - 1|\right]
\end{equation}
where $\|\cdot\|_2$ is the $L_2$ norm. 
Note that, since the volume of the whole $n$-sphere is larger than the volume of its  surface,  
 $\mathcal{L}_{sph}$ leads to a much more  compact space compared to $\mathcal{L}_{SR}$. Tab.~\ref{tab:sphere_comparisons} 
 quantitatively compares $\mathcal{L}_{sph}$  with $\mathcal{L}_{SR}$ and 
 shows  that a very compact space ($\mathcal{L}_{sph}$)  leads to a higher smoothness but with a low diversity. 
 This finding is qualitatively confirmed in
 Fig.~\ref{fig:sr-sph}. This comparison indicates that there exists a trade-off between the smoothness of the space and the  diversity of generated images. 

\begin{table}[!ht]
    \renewcommand{\tabcolsep}{3pt}
    \renewcommand{\arraystretch}{1}
    \small
	\centering
	\begin{tabular}{rrrrr}
	\toprule
     \textbf{Model} & \textbf{FID}$\downarrow$ & \textbf{LPIPS}$\uparrow$ & \textbf{PS}$\uparrow$ & \textbf{FRD}$\downarrow$ \\ \midrule
	 $\mathcal{L}_{SR}$ & \textbf{23.37} & \textbf{.337} & .504 & .837 \\
	 $\mathcal{L}_{sph}$ & 23.66 & .103 & \textbf{.897} & \textbf{.808}  \\ 
	\bottomrule
	\end{tabular}
	\vspace{-0.5em}
	\caption{A comparisons between $\mathcal{L}_{SR}$ and $\mathcal{L}_{sph}$ on a gender translation task using the CelebA-HQ dataset. 
	}\vspace{-4mm}
	\label{tab:sphere_comparisons}
\end{table} 

\begin{figure*}[!ht]	
	\renewcommand{\tabcolsep}{1pt}
	\renewcommand{\arraystretch}{0.2}
	\centering
	\footnotesize
	\begin{tabular}{cc}
	     (a) & \includegraphics[width=0.96\linewidth]{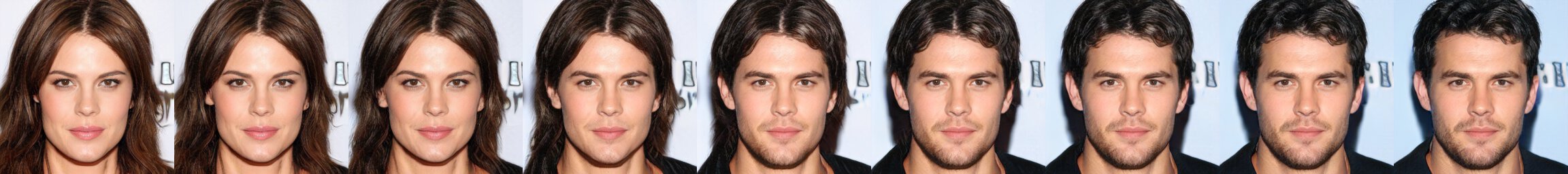} \\
	     (b) & \includegraphics[width=0.96\linewidth]{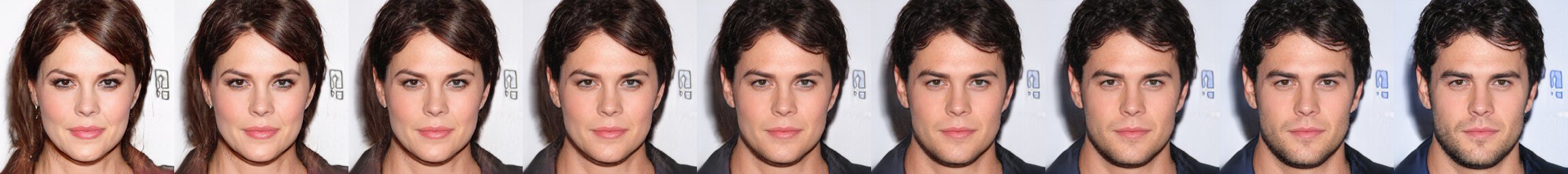} \\
	\end{tabular}
    \caption{Visual comparisons between (a) $\mathcal{L}_{SR}$ and (b) $\mathcal{L}_{sph}$.}
    \label{fig:sr-sph}
\end{figure*}

\subsection{A Space Visualization Experiment}

\begin{figure*}[ht]	
	\renewcommand{\tabcolsep}{1pt}
	\renewcommand{\arraystretch}{0.8}
	\centering
	\footnotesize
	\begin{tabular}{c}
		\includegraphics[width=0.96\linewidth]{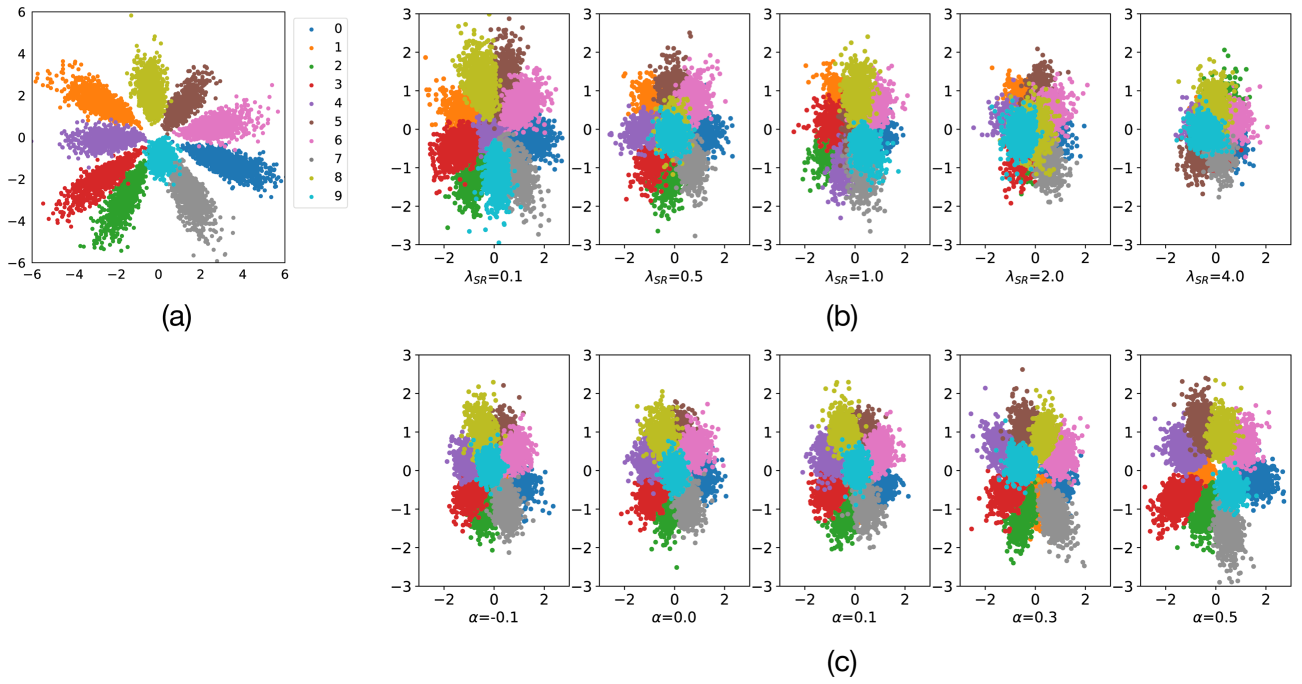} \\ 
	\end{tabular}
	\vspace{-1em}
	\caption{The distributions of style codes on a MNIST-based toy experiment. The original latent style space (a), using only $\mathcal{L}_{SR}$ with different loss weights $\lambda_{SR}$ (b), and using $\mathcal{L}_{SR}$ ($\lambda_{SR}=1.0$) and $\mathcal{L}_{tri}$ with different margin values $\alpha$ (c). 
	}
	\label{Fig:appendix_mnist_toy}
\end{figure*}

We perform an additional experiment on the MNIST dataset~\cite{gradient1998yann} to interpret the results of our model and directly visualize the distributions of style codes. In this experiment, we consider the categories of handwritten digits as ``styles" and we set the dimension of style codes to 2, such that they can be easily plotted in a two-dimensional coordinate system without reducing the representation dimensionality with non-linear projections (e.g. t-SNE). As shown in Fig.~\ref{Fig:appendix_mnist_toy} (a), the original style codes without using our proposed losses, is scattered  in a non-compact space, where there are many ``training gaps". Once we increase the weight of $\lambda_{SR}$, the style codes are pushed in a more compact space. However, the clusters (i.e., the domains) are highly entangled, as shown in Fig.~\ref{Fig:appendix_mnist_toy} (b). Conversely, the triplet  loss  alleviates this issue by separating the compacted clusters, as shown in Fig.~\ref{Fig:appendix_mnist_toy} (c). 

\begin{figure}[ht]	
	\renewcommand{\tabcolsep}{1pt}
	\renewcommand{\arraystretch}{0.8}
	\centering
	\footnotesize
	\begin{tabular}{cc}
		(a) & \includegraphics[width=0.96\linewidth]{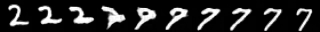} \\ 
		(b) & \includegraphics[width=0.96\linewidth]{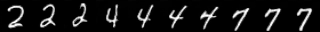} \\ 
		(c) & \includegraphics[width=0.96\linewidth]{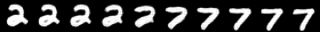} \\ 
	\end{tabular}
	\vspace{-1em}
	\caption{Interpolations results on MNIST between domain ``2"  and domain ``7".
 (a) Original space,  (b) Using only $\mathcal{L}_{SR}$ ($\lambda_{SR} = 1.0$). (c) Using   $\mathcal{L}_{SR}$ ($\lambda_{SR} = 1.0$) and $\mathcal{L}_{tri}$ ($\alpha=0.5$).}
	\label{Fig:appendix_mnist_toy_inter}
\end{figure}

Moreover, we select two clusters with large ``training gaps" (i.e., ``2" (green color) and ``7" (grey color)) in the original space Fig.~\ref{Fig:appendix_mnist_toy} (a). Fig.~\ref{Fig:appendix_mnist_toy_inter} (a) shows an example of  interpolation results between ``2" and ``7" with large ``training gaps", showing, as expected,  that the generated images contain artifacts. Fig.~\ref{Fig:appendix_mnist_toy_inter} (b) refers to the same interpolation  between ``2" and ``7" in the setting with $\lambda_{SR}=1.0$. It seems that, due to the  cluster  overlapping, the interpolation traverses another cluster (i.e., ``4") while moving from ``2" to ``7". Finally, the triplet  loss is able to disentangle the compact space, as shown in Fig.~\ref{Fig:appendix_mnist_toy_inter} (c), where no ``intruder'' is generated when interpolating between the two domains.

\section{PS Details}

The proposed PS score requires  a perceptual distance metric $\psi(\cdot, \cdot)$. We chose to use the LPIPS~\cite{zhang2018unreasonable} distance, which was shown to well align to human judgements.
However, although Zhang et al.~\cite{zhang2018unreasonable}   claim that LPIPS is a metric, its formulation is based on the squared Euclidean distance between deep learning features:
\begin{equation}
\label{eq:lpips_org}
    d(\pmb{x}_1, \pmb{x}_2) = \sum_{l}\frac{1}{H_lW_l}\sum_{h,w} w_l\|\pmb{y}_1^l - \pmb{y}_2^l\|_2^2
\end{equation}
where $\pmb{x}_1$ and $\pmb{x}_2$ are image patches, $\pmb{y}^l$ is a  feature extracted with a pretrained network $\mathcal{F}$ (e.g., AlexNet~\cite{krizhevsky2017imagenet}) using its $l$-th layer, and the weights $w_l$ are layer-specific weights
trained to mimic the human perception. Thus, Eq.(\ref{eq:lpips_org}) does not obey to the Triangle Inequality, which is necessary for a distance to be a metric. To avoid this problem, we re-train the $w_l$ weights using an Euclidean-distance formulation, which gives us a proper metric (called LPIPS* in the rest of this Appendix):
\begin{equation}
\label{eq:lpips_ours}
    d'(\pmb{x}_1, \pmb{x}_2) = \sum_{l}\frac{1}{H_lW_l}\sum_{h,w}w_l\|\pmb{y}_1^l - \pmb{y}_2^l\|_2.
\end{equation}
Following the original paper~\cite{zhang2018unreasonable}, the network $\mathcal{F}$ used in our paper is an AlexNet~\cite{krizhevsky2017imagenet} pre-trainted on ImageNet where a linear classifier (i.e., the the $w_l$ weights) is trained  to learn a human perception distance.

\begin{table}[!ht]
    \renewcommand{\tabcolsep}{3pt}
    \renewcommand{\arraystretch}{1}
    \small
	\centering
	\begin{tabular}{@{}llcccc cccc@{}}
	\toprule
	 \multirow{2}{*}{\textbf{Model}} & \textbf{Percep.} & \multicolumn{2}{c}{\textbf{PS}$\uparrow$} & \multicolumn{2}{c}{\textbf{LPIPS}$\downarrow$}   & \multicolumn{2}{c}{\textbf{PPL}$\downarrow$}  \\ \cmidrule(lr){3-4} \cmidrule(lr){5-6} \cmidrule(lr){7-8}
	  & \textbf{Distance} & Intra & Inter & Intra &  Inter & Intra &  Inter \\ \midrule
	  \multirow{2}{*}{\cite{choi2019stargan}}  & LPIPS & .877 & .670 & .005 & .012  & \multirow{2}{*}{19.21} & \multirow{2}{*}{57.19}\\ 
	  & LPIPS* & .545 & .359 & .061 & .107 \\ \cmidrule{2-8}
	 \multirow{2}{*}{Ours}  & LPIPS & .850 & .840 & .003 & .006 & \multirow{2}{*}{9.84} &  \multirow{2}{*}{22.78}  \\
	 & LPIPS* & .625 & .485 & .047 & .071 \\
	\bottomrule
	\end{tabular}
	\vspace{1pt}
	\caption{Comparing  different smoothness metrics. 
	We use  two different basic perceptual distances for all the metrics: the original LPIPS (Eq.\eqref{eq:lpips_org}) and the revised LPIPS* 
	(Eq.\eqref{eq:lpips_ours}).
 The LPIPS column refers to the diversity degree \cite{zhang2018unreasonable}. ``Intra" and ``Inter" refer to intra-domain  and inter-domain interpolations, respectively. 
}
	\label{tab:metric_comparisons}
\end{table}

\noindent\textbf{Comparison with other smoothness metrics.}
The smoothness of a latent style space can also be evaluated using LPIPS~\cite{zhang2018unreasonable} and the PPL~\cite{karras2019style} scores. In  ideally smooth interpolations, the perceptual distance (LPIPS) between two neighbouring interpolations should be as low as possible (i.e., high similarity). 
Similarly, the PPL should be as low as possible to indicate the smoothness of the space. Note that when the model exhibits a mode collapse problem, we can have PPL=0 (or LPIPS=0).
Despite this, we compare the LPIPS, PPL and PS scores on an additional experiment, where we randomly use 
both intra and inter-domain
interpolation lines. For each interpolation line we  generate 20  images.
Tab.~\ref{tab:metric_comparisons} shows that: (1) the higher the PS score, usually the lower the LPIPS and the PPL score; (2) our PS metric based on LPIPS* is  more consistent with the LPIPS and the PPL with respect to the smoothness degree. 
Moreover, our PS metric is more interpretable, as it ranges between 0 and 1, while the alternatives range in $[0, \infty]$.

\begin{table}[!ht]
    \renewcommand{\tabcolsep}{3pt}
    \renewcommand{\arraystretch}{1}
    \small
	\centering
	\begin{tabularx}{\columnwidth}{@{}Xrrrr@{}}
	\toprule
	\multirow{2}{*}{\textbf{Percepual Distance}} & \multicolumn{4}{c}{\textbf{Num. of Interpolation}} \\ \cmidrule{2-5}
	 & 10 & 20 & 50 & 100 \\ \midrule 
    PPL & 120.63 & 457.53 & 2122.33 & 6369.93 \\
    LPIPS & 0.133 & 0.106 & 0.066 & 0.042 \\
    LPIPS* & 1.150 & 1.424 & 1.723 & 1.908 \\
	\bottomrule
	\end{tabularx}
	\caption{The sum of the perceptual distances along the same interpolation lines averaged over all the generated images. This table shows the linearity of various perceptual distance metrics.}
	\label{tab:linearity}
\end{table} 

\noindent\textbf{Number of Interpolations.} We also compute the robustness of the different metrics on a high number of interpolations in Tab.~\ref{tab:linearity},  where  we use the same start-end  style codes for all the metrics.
Tab.~\ref{tab:linearity} shows that PPL is not a linear metric and it is sensitive to the interpolation step size (i.e., the smaller the interpolation step size, the larger the PPL score). Similarly, LPIPS is also not a linear metric and it tends to decrease when the number of interpolations increase. Conversely, the proposed PS score is consistent, it satisfies the triangle inequality and its behaviour is more linear.

\begin{table}[!ht]
    \renewcommand{\tabcolsep}{3pt}
    \renewcommand{\arraystretch}{1}
    \small
	\centering
	\begin{tabularx}{\columnwidth}{@{}Xc rr rr rr@{}}
	\toprule
	\multirow{2}{*}{\textbf{Model}} & \multirow{2}{*}{\textbf{Interpolation}} & \multicolumn{2}{c}{\textbf{PS}$\uparrow$} & \multicolumn{2}{c}{\textbf{LPIPS}$\downarrow$}   & \multicolumn{2}{c}{\textbf{PPL}$\downarrow$}  \\ 
	\cmidrule(lr){3-4} \cmidrule(lr){5-6} \cmidrule(lr){7-8} 
	 & & Intra & Inter & Intra &  Inter & Intra &  Inter \\ \midrule
	 \cite{choi2019stargan} & \multirow{2}{*}{Lerp} &  .545 & .359 & .061 & .107 & 19.21 & 57.19 \\ 
	 Ours &  &  \textbf{.625} & \textbf{.485} & \textbf{.047} & \textbf{.071} & \textbf{9.84} & \textbf{22.78} \\
	\midrule\midrule
	  \cite{choi2019stargan} & \multirow{2}{*}{Slerp} & .531 & .336 & .065 & .120 & 19.69 & 64.81 \\ 
	 Ours &  & \textbf{.607} & \textbf{.404} & \textbf{.049} & \textbf{.083} & \textbf{10.53} & \textbf{26.17} \\
	\bottomrule
	\end{tabularx}
	\caption{Different interpolation strategies. Both StarGAN v2~\cite{choi2019stargan} and the our method achieve a better performance with ``lerp".}
	\label{tab:slerp}
\end{table}

\noindent\textbf{Interpolation Strategies.} Finally, we test the robustness of the PS score with respect to two different interpolation strategies (i.e., \textit{lerp} and \textit{slerp} \cite{karras2019style}). As shown in Tab.~\ref{tab:slerp}, both  our methdod and StarGAN v2~\cite{choi2019stargan} achieve a slightly better result when using the linear interpolation (\textit{lerp}), which  indicates the linearity of the  style space.

\section{Face Recognition Distance}

\begin{figure*}[!ht]	
	\renewcommand{\tabcolsep}{1pt}
	\renewcommand{\arraystretch}{0.8}
	\newcommand{\sizea}{0.124\linewidth}
	\centering
	\begin{tabular}{c|ccccccc}
	     Input & & & \multicolumn{2}{c}{Synthesized Images} \\
		\includegraphics[width=\sizea]{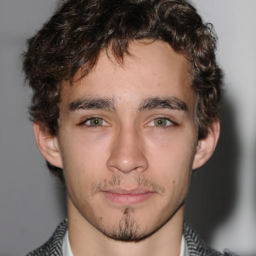} &
		\includegraphics[width=\sizea,cframe=red 1pt]{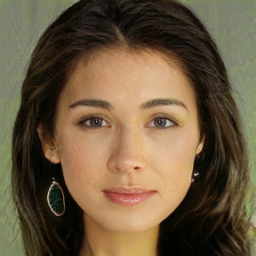} &
		\includegraphics[width=\sizea,cframe=green 1pt]{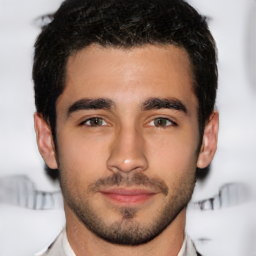} & 
		\includegraphics[width=\sizea,cframe=green 1pt]{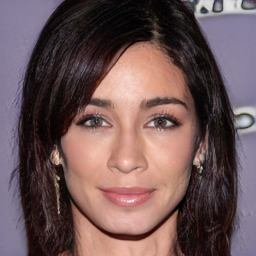} &
		\includegraphics[width=\sizea,cframe=red 1pt]{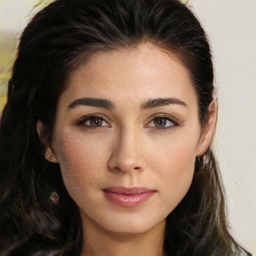} & 
		\includegraphics[width=\sizea,cframe=green 1pt]{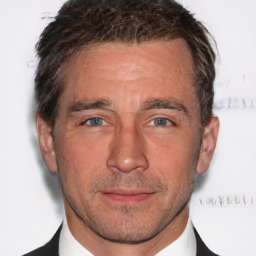} & 
		\includegraphics[width=\sizea,cframe=red 1pt]{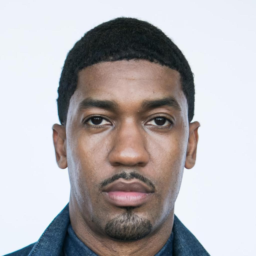}\\
		& 1.276 & 0.963 & 1.041 & 1.212  & 0.906 & 1.419\\
	\end{tabular}
    \caption{The synthesized images with ``green” bounding box are with lower FRD scores, in which identity features are preserved better. However, FID and IS metrics are not aware of identity preserving. 
    }
    \label{fig:reid-demo}
\end{figure*}

\begin{figure*}[!ht]	
	\renewcommand{\tabcolsep}{1pt}
	\renewcommand{\arraystretch}{0.8}
	\newcommand{\sizea}{\linewidth}
	\centering
	\footnotesize
	\begin{tabular}{c}
		\includegraphics[width=\sizea]{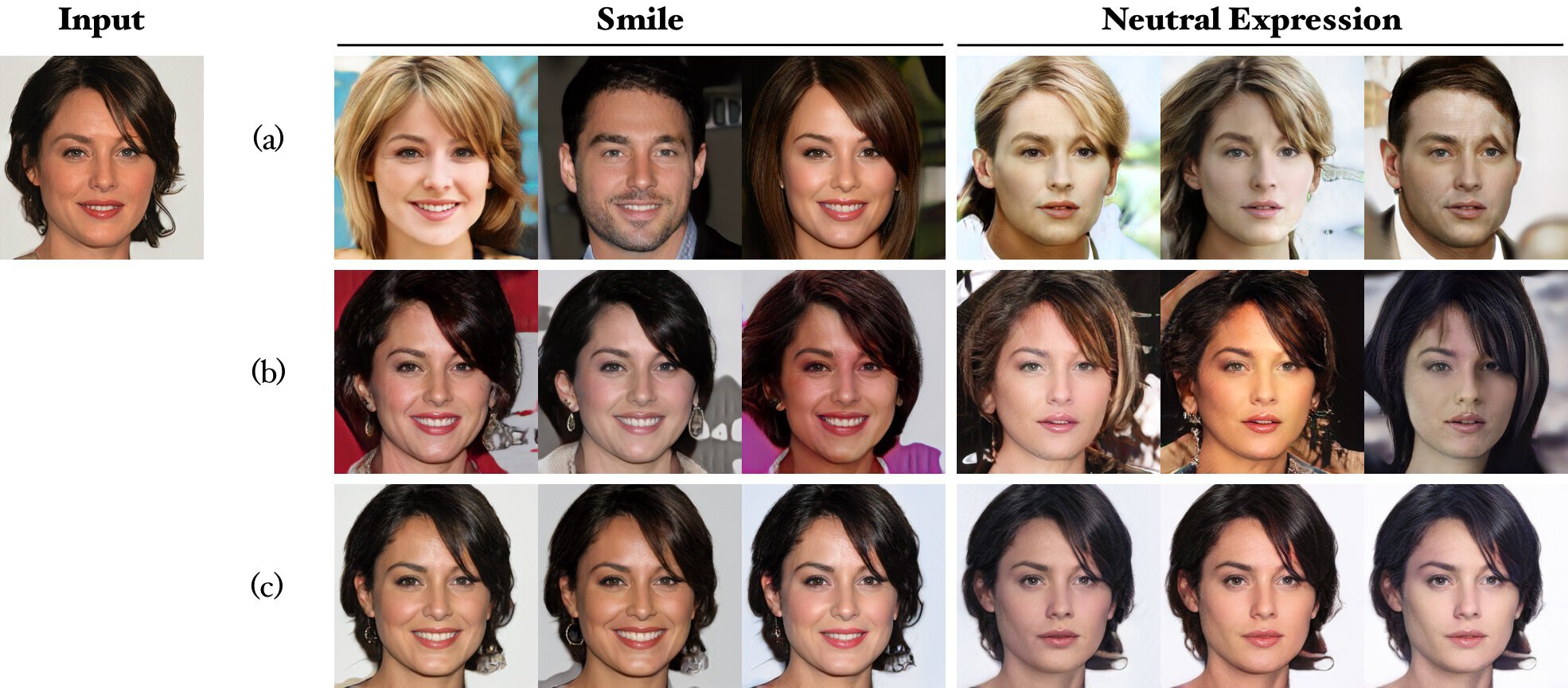}
	\end{tabular}
    \caption{A visual comparison for the  smile translation task on the CelebA-HQ dataset. (a) StarGAN v2~\cite{choi2019stargan}, (b) our proposed method without $\mathcal{L}_{cont}$ and (c) our proposed method with $\mathcal{L}_{cont}$.  This comparison shows that a smooth  style space can better  preserve the person identity. Moreover, using $\mathcal{L}_{cont}$ significantly boosts the  the input identity preservation. 
    }
    \label{fig:smile-ablation}
\end{figure*}

Fig.~\ref{fig:reid-demo} shows an example of face translation,  which indicates the crucial issue of identity preservation. For example, an arbitrary female face can be realistic for a discriminator, but if the original-person identity is completely lost, this is not the desired output of a gender translation. Fig.~\ref{fig:smile-ablation} shows a comparisons based on a smile translations task on the CelebA-HQ dataset, which further shows the importance of the identity preservation. The StarGAN generated images frequently loose the identity of the source images, while ours  do not. Moreover, we see that $\mathcal{L}_{cont}$ is very important both for the identity and the background preservation.

\section{LPIPS for Diversity}
The state of the art models are often evaluated through the LPIPS distance. Usually, for each input, different target styles are randomly sampled. Then, the LPIPS is computed on all the generated outputs to model the diversity (also called multi-modality) of the generated images. 
However, a high LPIPS distance is not always desirable. 
For example, a high LPIPS value can be produced also when:
\begin{itemize}
    \item The generated images do not always look real (e.g. the images with artifacts shown in the first row of \Cref{fig:more_celebahq_comparisons}).
    \item The domain-independent part of the image is not preserved. For example, when the background appearance has drastically changed (e.g.,  \Cref{fig:smile-ablation} (a)) or when the person-identity is not preserved (e.g., \Cref{fig:smile-ablation} (a) and \Cref{fig:smile-ablation} (b)).
\end{itemize}
For these reasons, we believe that in an MMUIT task,  LPIPS scores should be taken with a pinch of salt, especially when the model is not good enough to preserve the domain-independent part of the source image.

\section{Additional Details}
\subsection{Datasets} Following  StarGAN v2~\cite{choi2019stargan}, we use the CelebA-HQ~\cite{karras2017progressive} and the AFHQ~\cite{choi2019stargan} dataset. CelebA-HQ is a high-quality version of the CelebA~\cite{liu2015deep} dataset, consisting of 30,000 images with a 1024$\times$1024 resolution. We randomly select 2,000 images for testing and we use all the remaining images for training. Differently from StarGAN v2, we also test the smile and the age attributes. AFHQ consists of 15,000 high-quality images at 512$\times$512 resolution. The dataset includes three domains (cat, dog, and wildlife),  with 5,000 images each. We select 500 images as the test set for each domain and we use all the remaining images for training. AFHQ and CelebA-HQ are tested at a 256$\times$256 resolution (note that we use a 128$\times$128 resolution in the comparisons with TUNIT~\cite{baek2020tunit}). 
In this Appendix we also used the low-resolution MNIST~\cite{lecun1998gradient} dataset, which consists of 60,000 training samples and 10,000 testing samples with a 32$\times$32 resolution.

\subsection{Compared Methods} 
We use the official released codes for all the compared methods, including StarGAN v2~\cite{choi2019stargan}\footnote[1]{https://github.com/clovaai/stargan-v2}, HomoGAN~\cite{chen2019homomorphic}\footnote[2]{https://github.com/yingcong/HomoInterpGAN}, InterFaceGAN~\cite{shen2020interpreting}\footnote[3]{https://github.com/genforce/interfacegan} and TUNIT~\cite{baek2020tunit}\footnote[4]{https://github.com/clovaai/tunit}. 
In the main paper (Sec. 4.2) we show how our proposed losses are combined with (i.e., simply added to) the  
 StarGAN v2 losses. Similarly, in case  of TUNIT, we use all the original losses of \cite{baek2020tunit} ($\mathcal{L}_{tunit}$) and we add 
 $\mathcal{L}_{SR}$ and $\mathcal{L}_{tri}$ (without using our content loss), which leads to: $\mathcal{L}_{tunit}+\mathcal{L}_{SR}+\mathcal{L}_{tri}$. 

 InterFaceGAN~\cite{shen2020interpreting} is not a I2I translation model, and
 there is no separation between the ``content'' and the ``style'' representations. Moreover, this method linearly interpolates codes on a StyleGAN~\cite{karras2019style} pre-trained  semantic space. Thus, it is not easy to fairly compare MMUIT models with InterFaceGAN.
In our paper, when we compare MMUIT models with InterFaceGAN, we start from a StyleGAN generated image $\pmb{x}$ and we modify its semantics by generating two new images $\pmb{x}' = G(\pmb{z} + -3\pmb{n}))$ and $\pmb{x}'' = G(\pmb{z} + 3\pmb{n}))$, where $\pmb{n}$ is the unit normal vector defining a domain-separation  hyperplane  (e.g. smile vs non-smile) learned by InterFaceGAN. In the semantic space of smile, $\pmb{x}'$ is an image with \emph{no smile}, while $\pmb{x}''$ an image with \emph{more smile}.
These two randomly images are then used as the reference images for the encoders of each compared model (including ours) to generate the style codes. 
Note that, using StyleGAN based reference images, most likely favours InterFaceGAN with respect to all the other compared methods.

\section{More Experiments}
\begin{figure*}[!ht]	
	\renewcommand{\tabcolsep}{1pt}
	\renewcommand{\arraystretch}{0.8}
	\newcommand{\sizea}{\linewidth}
	\centering
	\footnotesize
	\begin{tabular}{c}
		\includegraphics[width=\sizea]{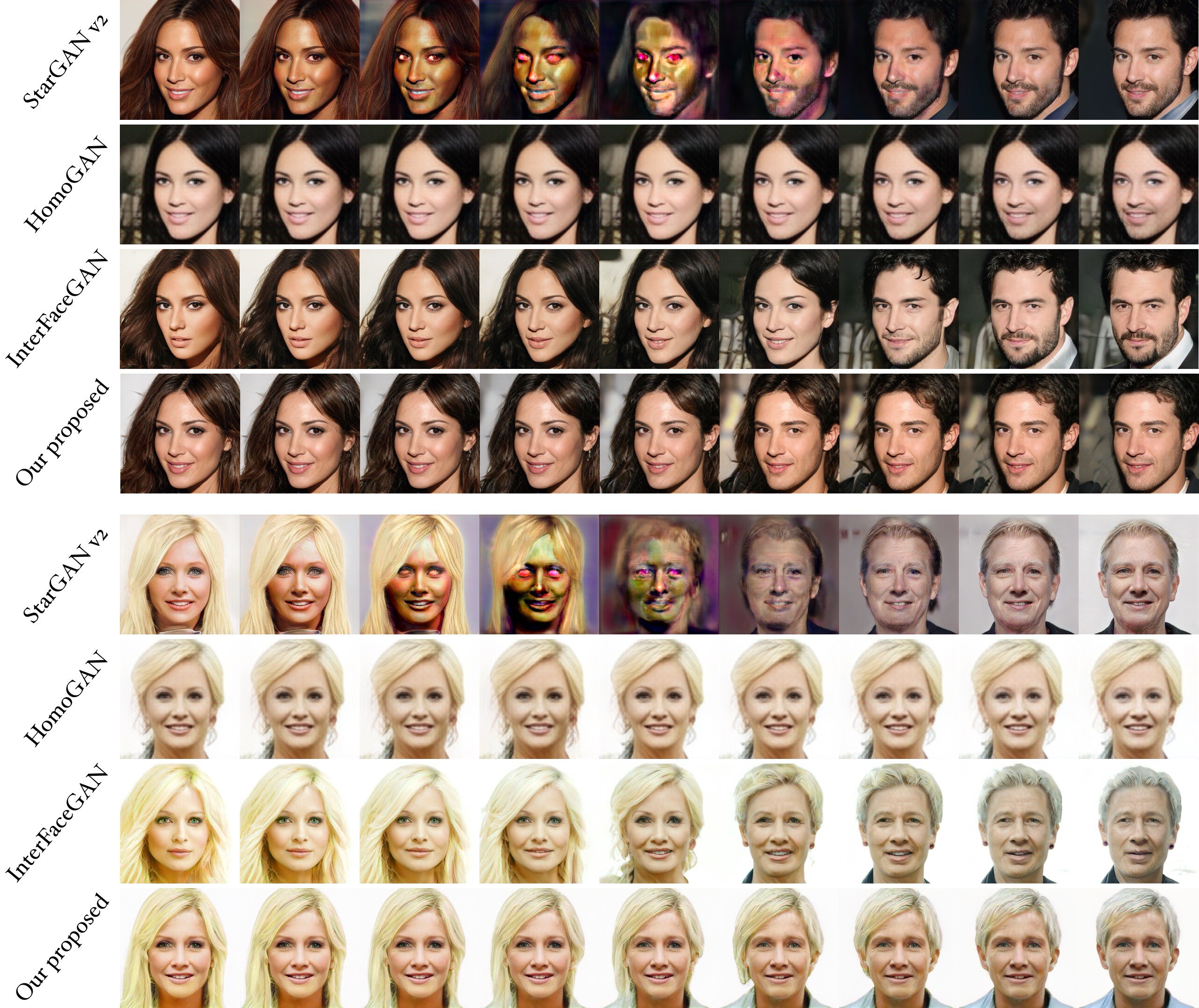}
	\end{tabular}
    \caption{Additional comparisons between StarGAN v2~\cite{choi2019stargan}, HomoGAN~\cite{chen2019homomorphic}, InterFaceGAN~\cite{shen2020interpreting} and our proposed method on a gender translation task on the CelebA-HQ dataset~\cite{karras2017progressive}. 
    }
    \label{fig:more_celebahq_comparisons}
\end{figure*}

\begin{figure*}[!ht]	
	\renewcommand{\tabcolsep}{1pt}
	\renewcommand{\arraystretch}{0.8}
	\newcommand{\sizea}{\linewidth}
	\centering
	\footnotesize
	\begin{tabular}{c}
		\includegraphics[width=\sizea]{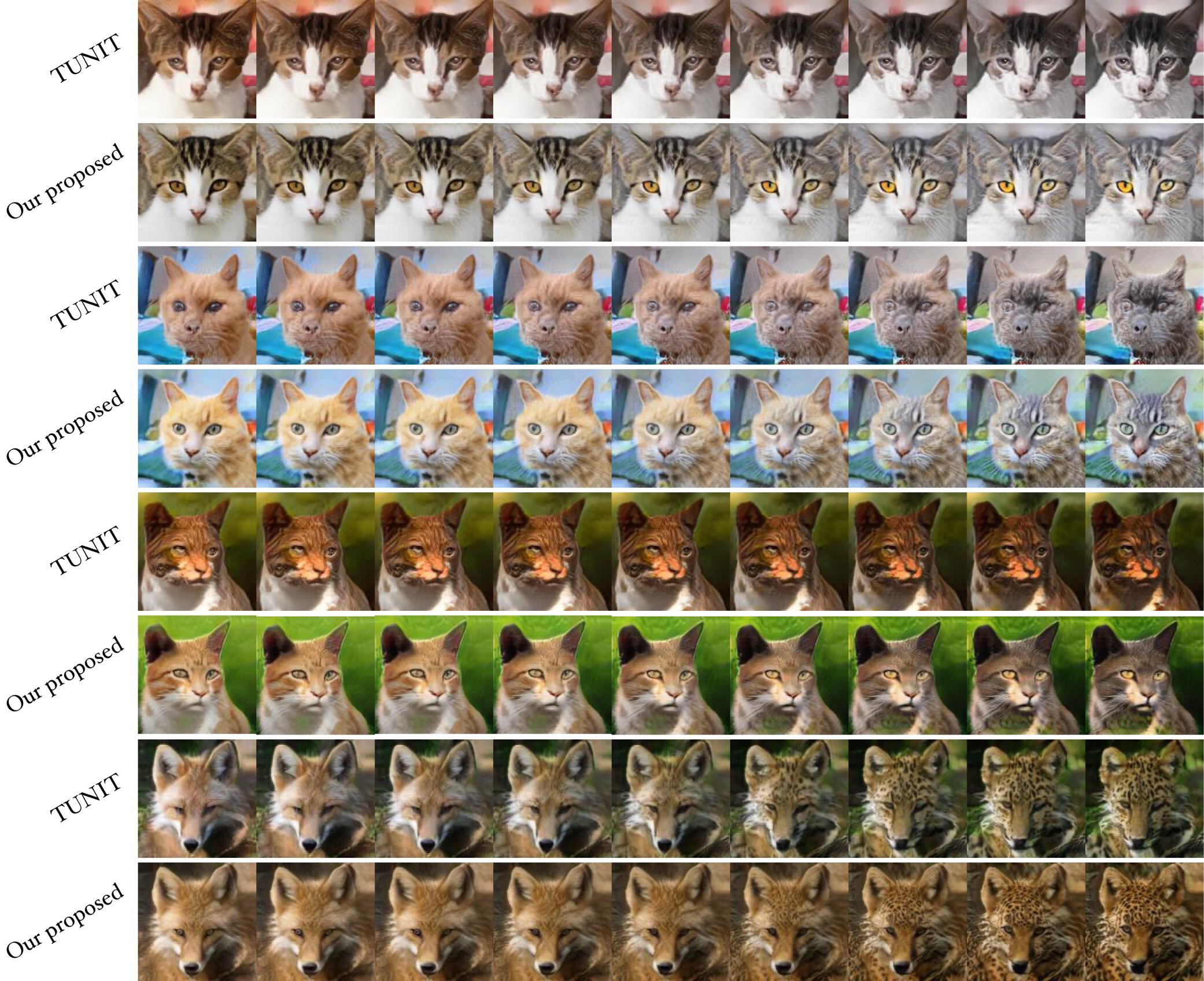}
	\end{tabular}
    \caption{An additional comparison between  TUNIT~\cite{baek2020tunit} and our proposed method on a truly unsupervised image-to-image translation task using the AFHQ dataset~\cite{choi2019stargan} (domain-level annotations are not provided). 
    }
    \label{fig:more_tunit_comparisons}
\end{figure*}

\begin{figure*}[!ht]	
	\renewcommand{\tabcolsep}{1pt}
	\renewcommand{\arraystretch}{0.8}
	\newcommand{\sizea}{\linewidth}
	\centering
	\footnotesize
	\begin{tabular}{c}
		\includegraphics[width=\sizea]{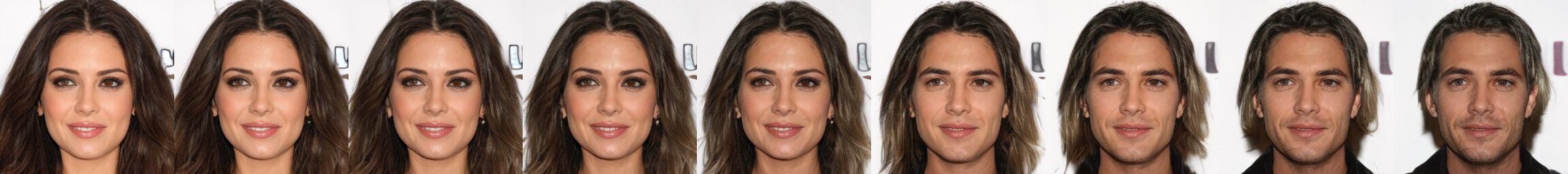} \\
		\includegraphics[width=\sizea]{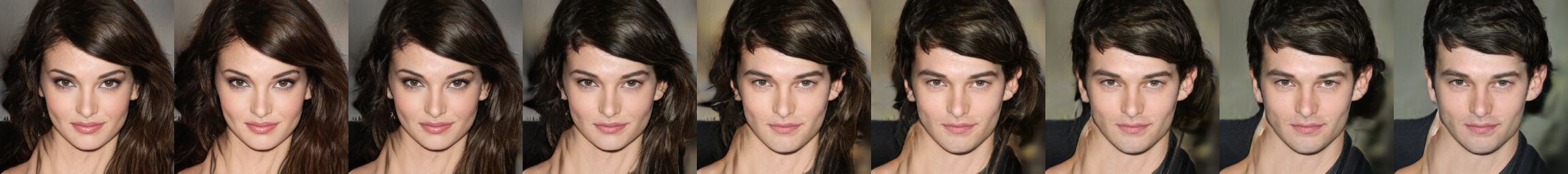} \\
		\includegraphics[width=\sizea]{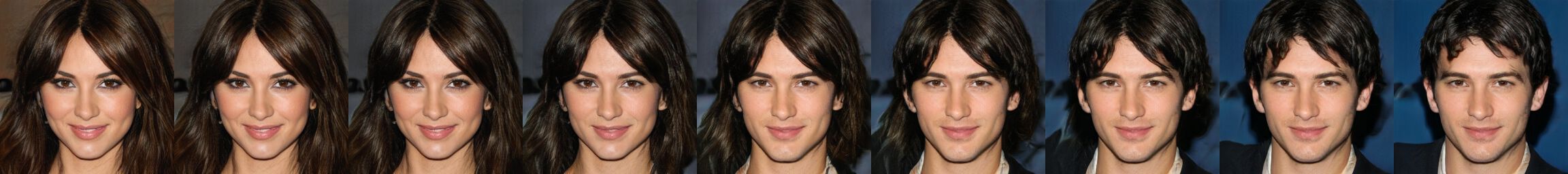} \\
		\includegraphics[width=\sizea]{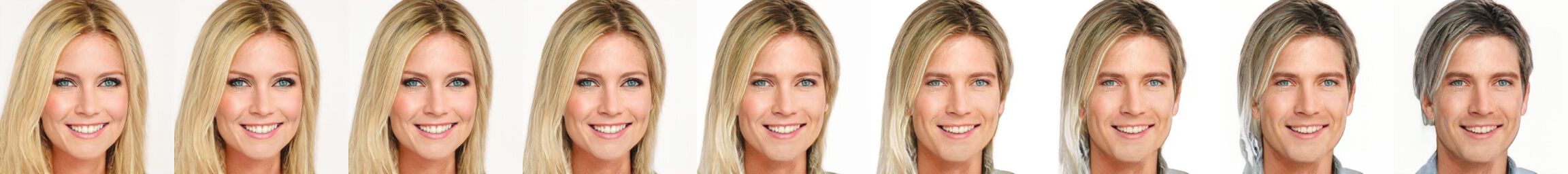} \\
		\includegraphics[width=\sizea]{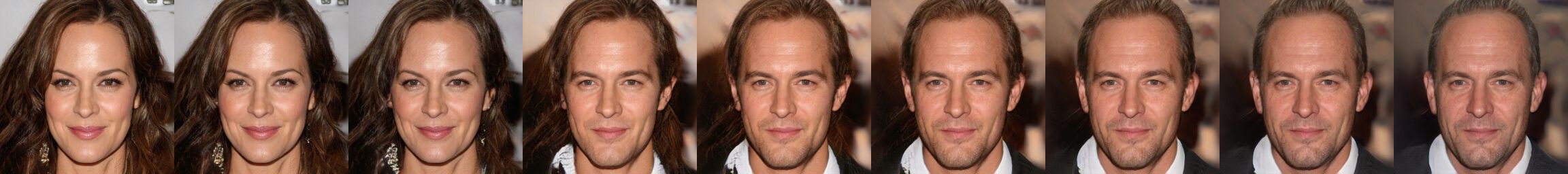} \\
		\includegraphics[width=\sizea]{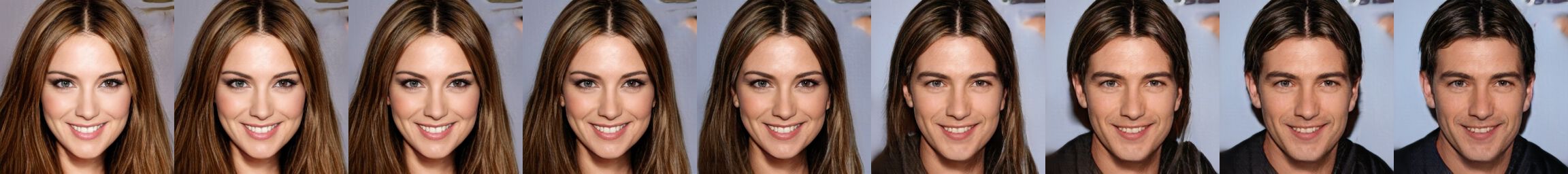} \\
		\includegraphics[width=\sizea]{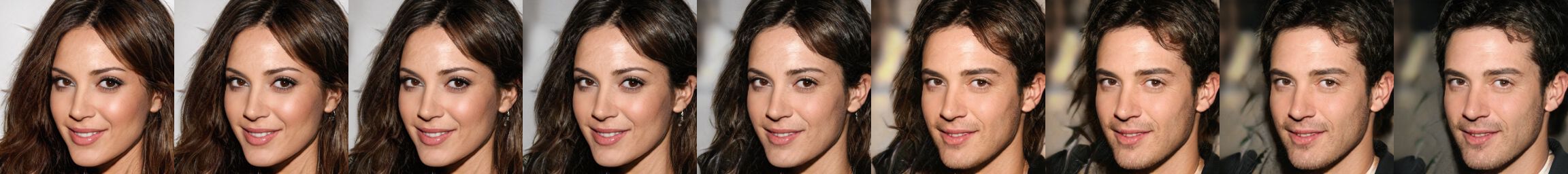} \\
		\includegraphics[width=\sizea]{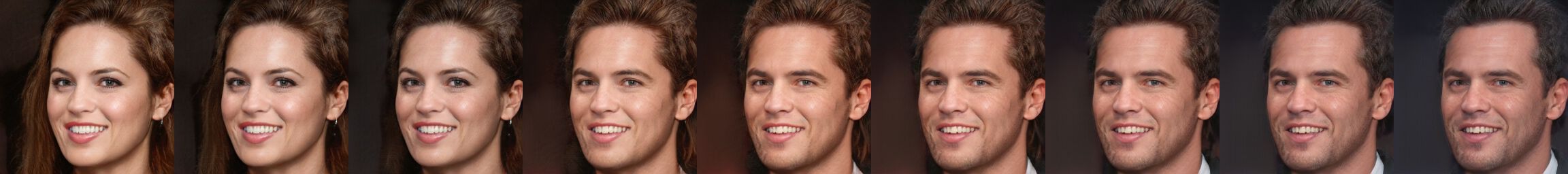} \\
	\end{tabular}
    \caption{More examples of gender translation on the CelebA-HQ dataset~\cite{karras2017progressive}. 
    }
    \label{fig:more_gender}
\end{figure*}

\begin{figure*}[!ht]	
	\renewcommand{\tabcolsep}{1pt}
	\renewcommand{\arraystretch}{0.8}
	\newcommand{\sizea}{\linewidth}
	\centering
	\footnotesize
	\begin{tabular}{c}
		\includegraphics[width=\sizea]{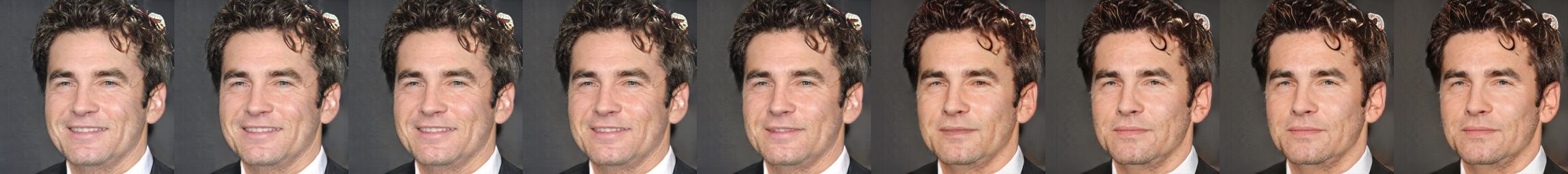} \\
		\includegraphics[width=\sizea]{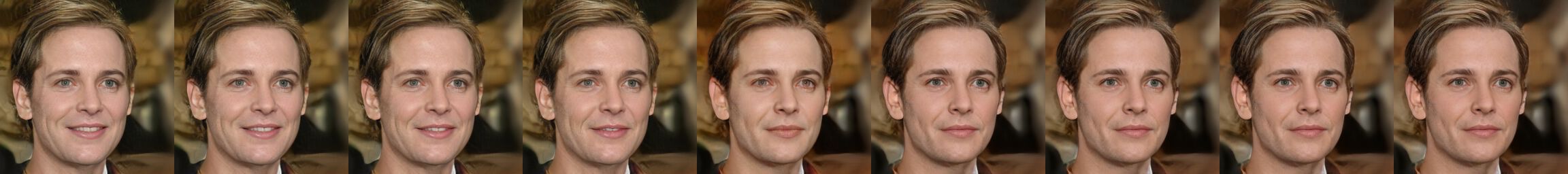} \\
		\includegraphics[width=\sizea]{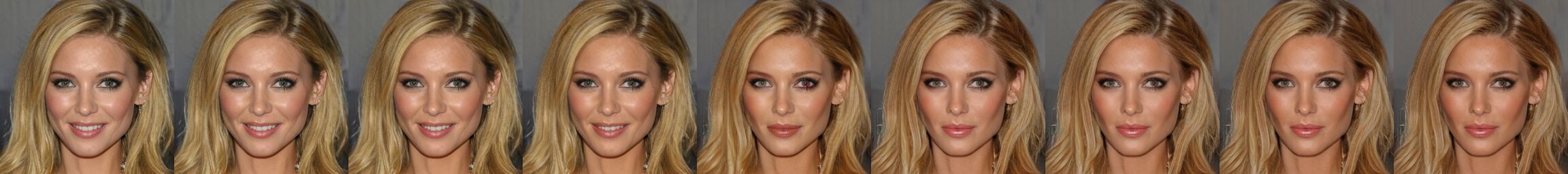} \\
		\includegraphics[width=\sizea]{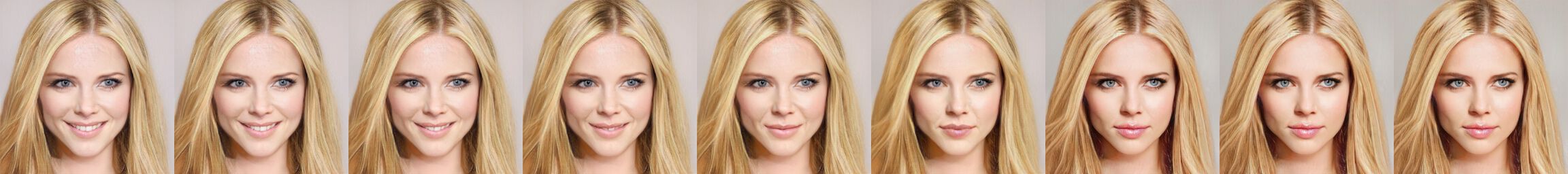} \\
		\includegraphics[width=\sizea]{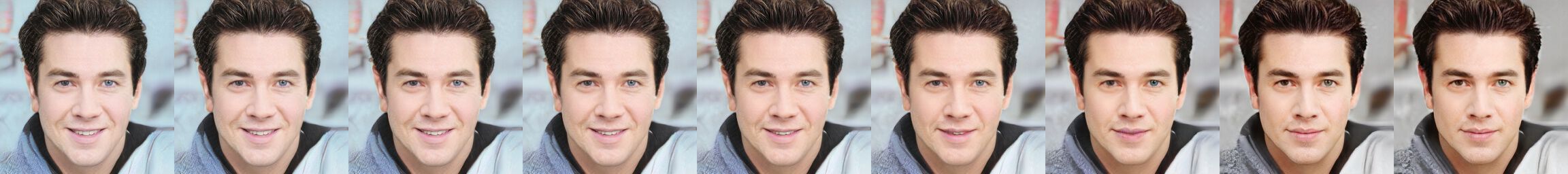} \\
		\includegraphics[width=\sizea]{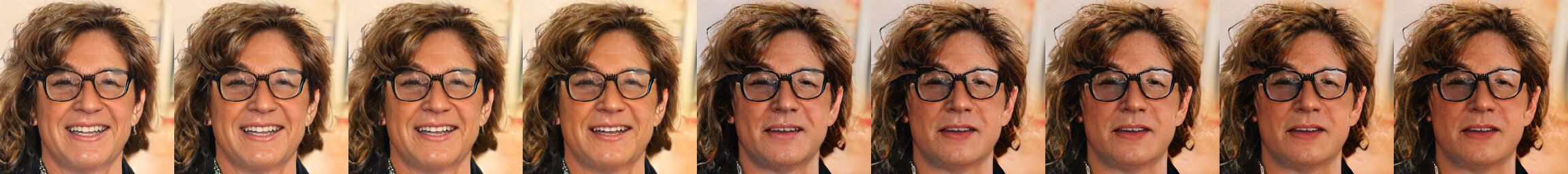} \\
		\includegraphics[width=\sizea]{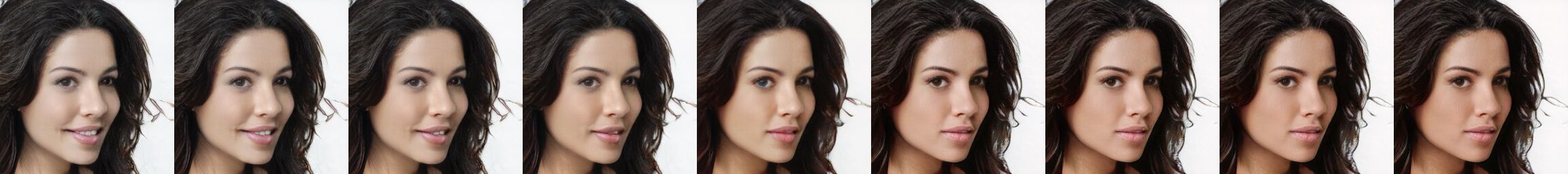} \\
		\includegraphics[width=\sizea]{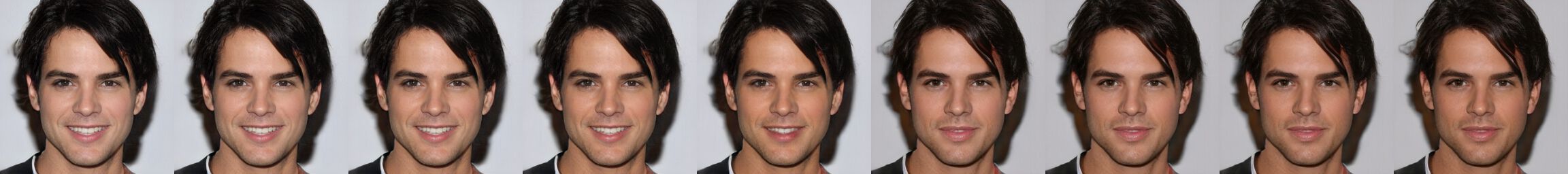} \\
	\end{tabular}
    \caption{More examples of smile translations on the CelebA-HQ dataset~\cite{karras2017progressive}. 
    }
    \label{fig:more_smile}
\end{figure*}

\begin{figure*}[!ht]	
	\renewcommand{\tabcolsep}{1pt}
	\renewcommand{\arraystretch}{0.8}
	\newcommand{\sizea}{\linewidth}
	\centering
	\footnotesize
	\begin{tabular}{c}
		\includegraphics[width=\sizea]{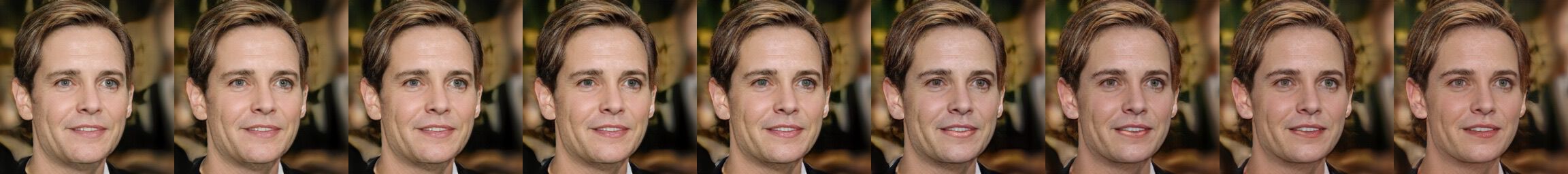} \\
		\includegraphics[width=\sizea]{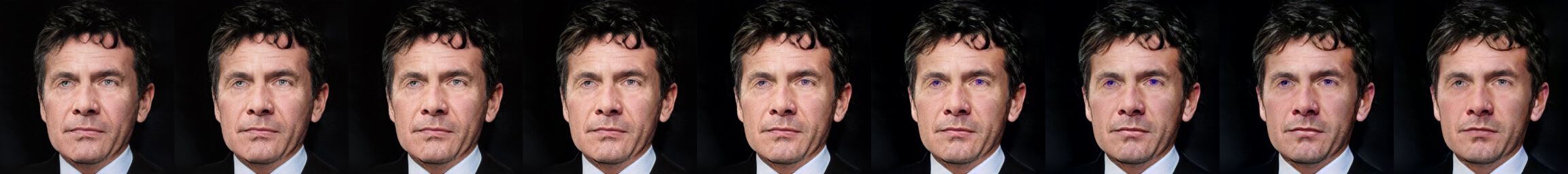} \\
		\includegraphics[width=\sizea]{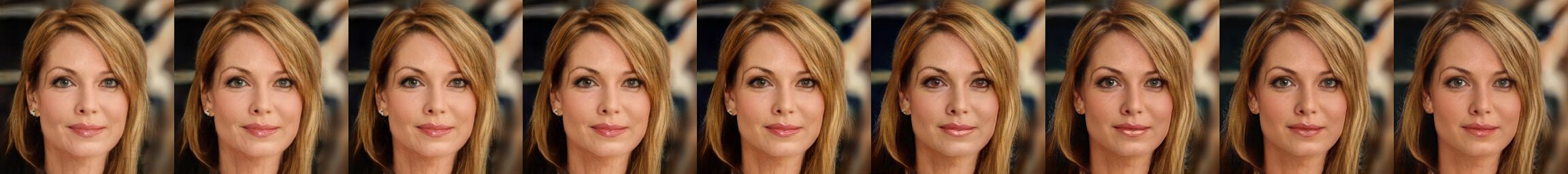} \\
		\includegraphics[width=\sizea]{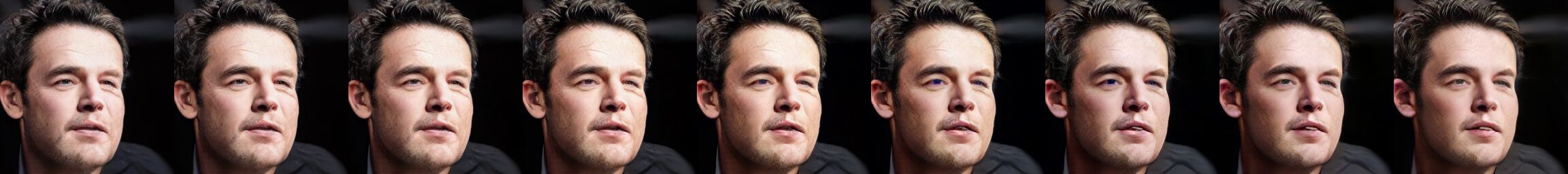} \\
		\includegraphics[width=\sizea]{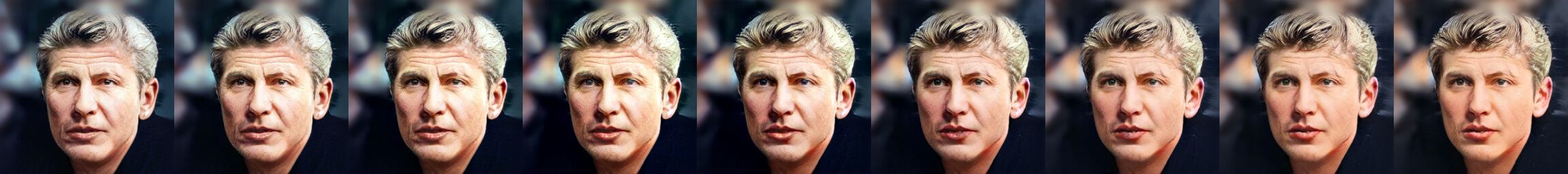} \\
		\includegraphics[width=\sizea]{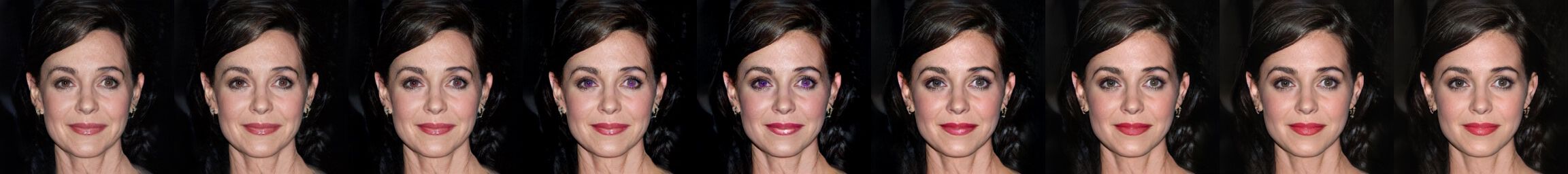} \\
		\includegraphics[width=\sizea]{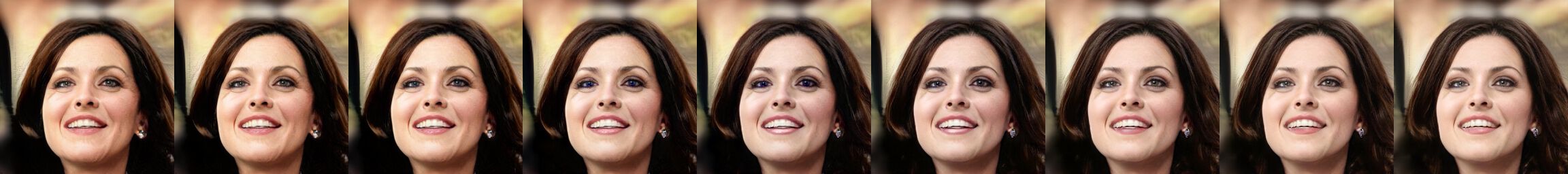} \\
		\includegraphics[width=\sizea]{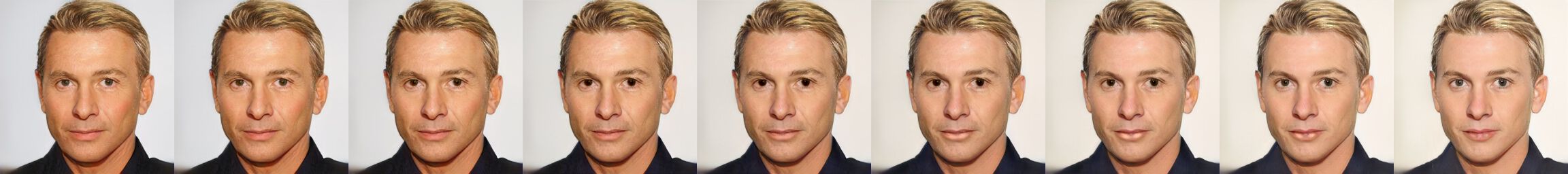} \\
	\end{tabular}
    \caption{More examples of age translations on the CelebA-HQ dataset~\cite{karras2017progressive}. 
    }
    \label{fig:more_age}
\end{figure*}

\begin{figure*}[!ht]	
	\renewcommand{\tabcolsep}{1pt}
	\renewcommand{\arraystretch}{0.8}
	\newcommand{\sizea}{\linewidth}
	\centering
	\footnotesize
	\begin{tabular}{c}
		\includegraphics[width=\sizea]{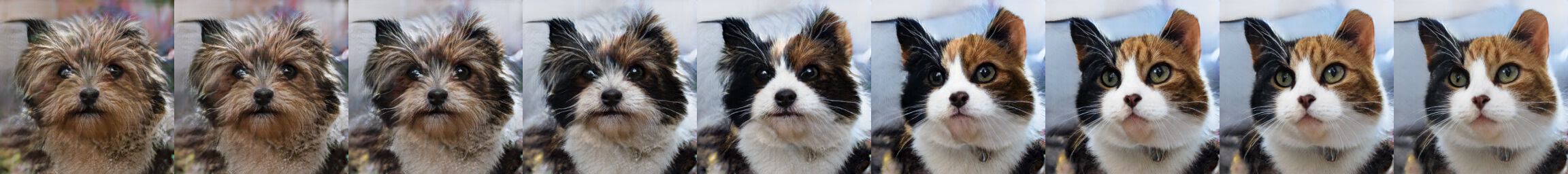} \\
		\includegraphics[width=\sizea]{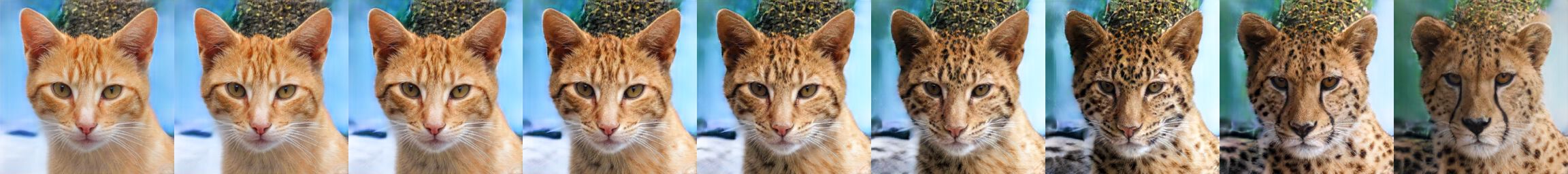} \\
		\includegraphics[width=\sizea]{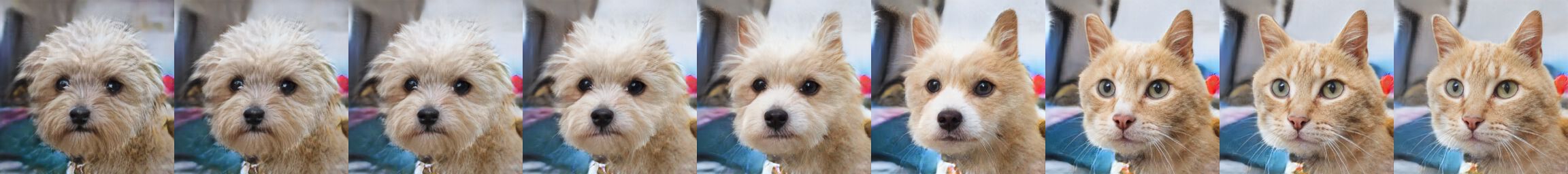} \\
		\includegraphics[width=\sizea]{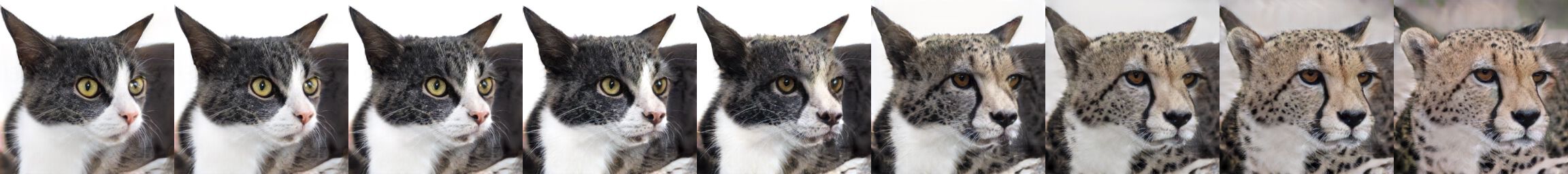} \\
		\includegraphics[width=\sizea]{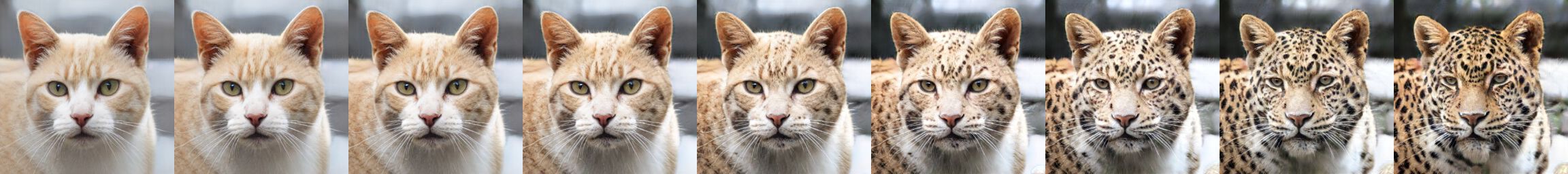} \\
		\includegraphics[width=\sizea]{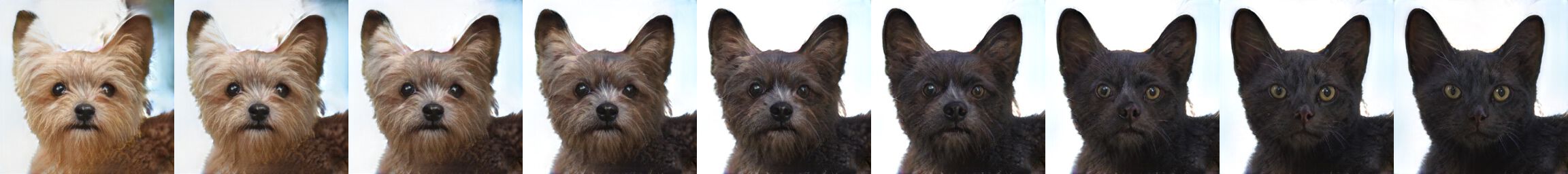} \\
		\includegraphics[width=\sizea]{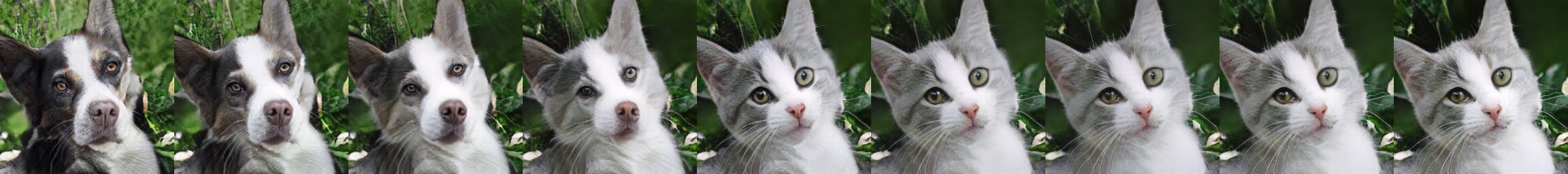} \\
		\includegraphics[width=\sizea]{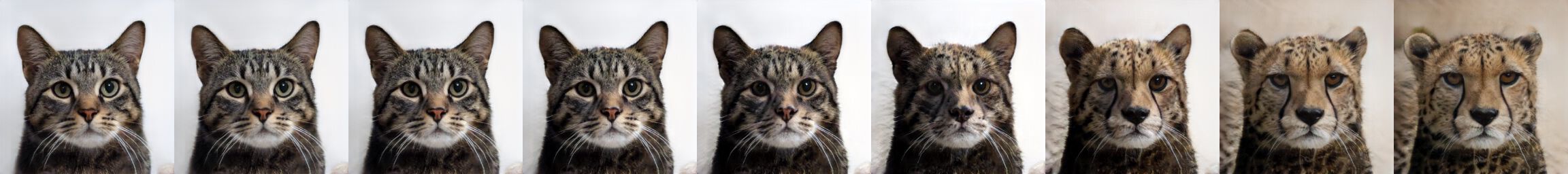} \\
	\end{tabular}
    \caption{More examples of animal translations on the AFHQ dataset~\cite{choi2019stargan}. 
    }
    \label{fig:more_afhq_cat}
\end{figure*}

More visual comparisons with StarGAN v2~\cite{choi2019stargan}, HomoGAN~\cite{chen2019homomorphic}, InterFaceGAN~\cite{shen2020interpreting} and TUNIT~\cite{baek2020tunit} are shown in Fig.~\ref{fig:more_celebahq_comparisons}-\ref{fig:more_tunit_comparisons}.
Fig.~\ref{fig:more_gender}-\ref{fig:more_age} show more visual results of gender, smile and age translations on the CelebA-HQ dataset. Fig.~\ref{fig:more_afhq_cat} shows more visual results of animal translations on the AFHQ dataset. 

\begin{figure*}[!ht]	
	\renewcommand{\tabcolsep}{1pt}
	\renewcommand{\arraystretch}{0.8}
	\newcommand{\sizea}{\linewidth}
	\centering
	\footnotesize
	\begin{tabular}{c}
		\includegraphics[width=\sizea]{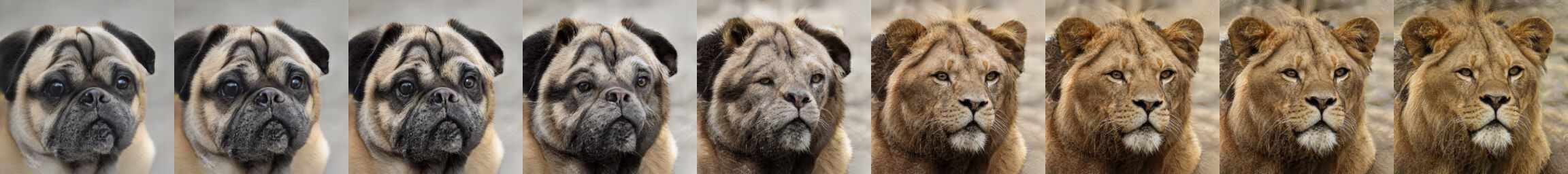} \\
		\includegraphics[width=\sizea]{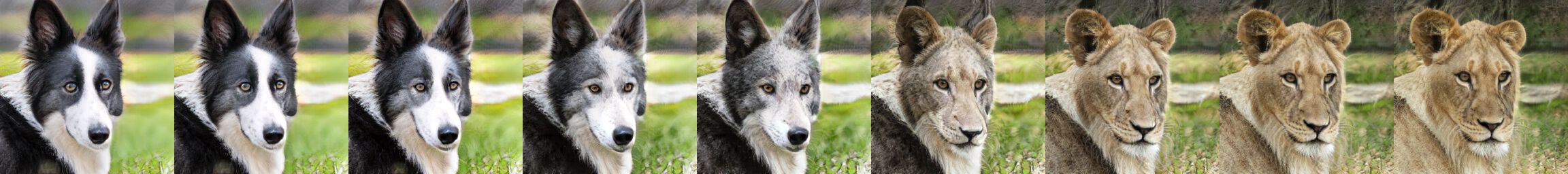} \\
		\includegraphics[width=\sizea]{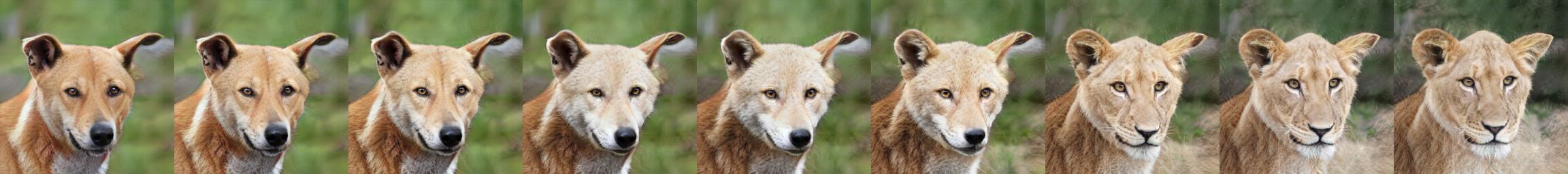} \\
		\includegraphics[width=\sizea]{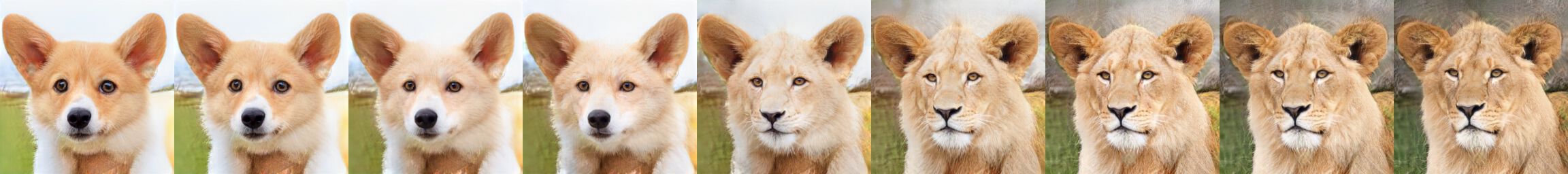} \\
		\includegraphics[width=\sizea]{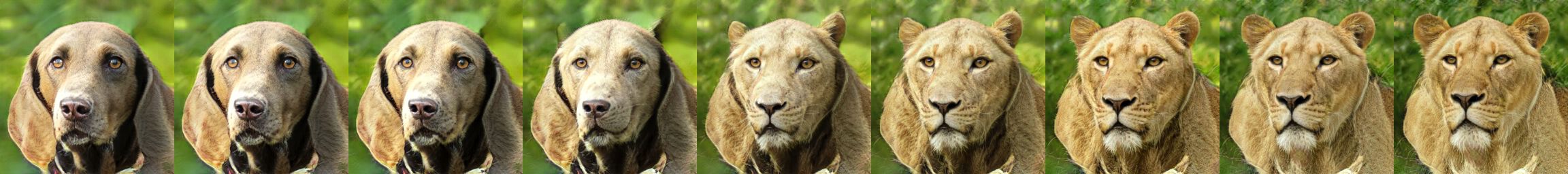} \\
		\includegraphics[width=\sizea]{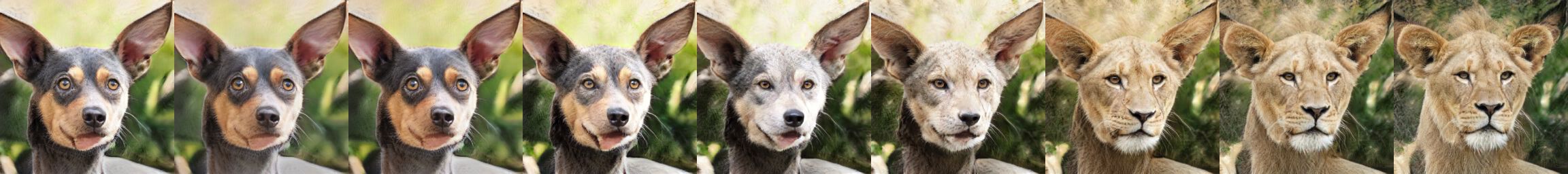} \\
		\includegraphics[width=\sizea]{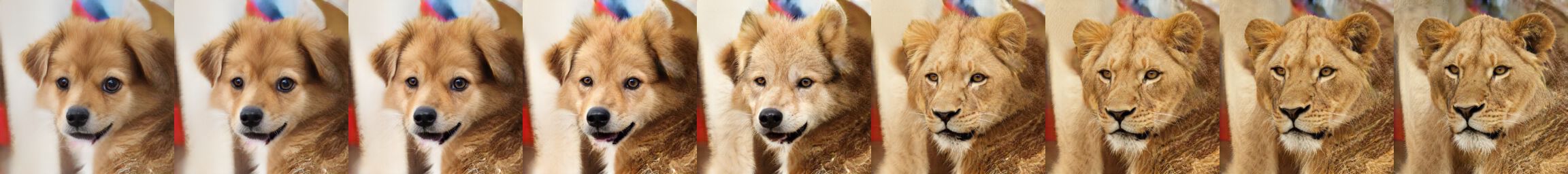} \\
		\includegraphics[width=\sizea]{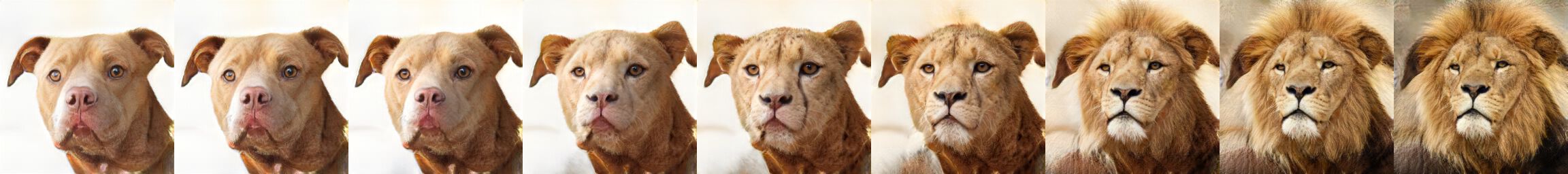} \\
	\end{tabular}
    \caption{More examples of animal translations on the AFHQ dataset~\cite{choi2019stargan}. 
    }
    \label{fig:more_afhq_dog}
\end{figure*}

\begin{figure*}[!ht]	
	\renewcommand{\tabcolsep}{1pt}
	\renewcommand{\arraystretch}{0.8}
	\newcommand{\sizea}{\linewidth}
	\centering
	\footnotesize
	\begin{tabular}{c}
		\includegraphics[width=\sizea]{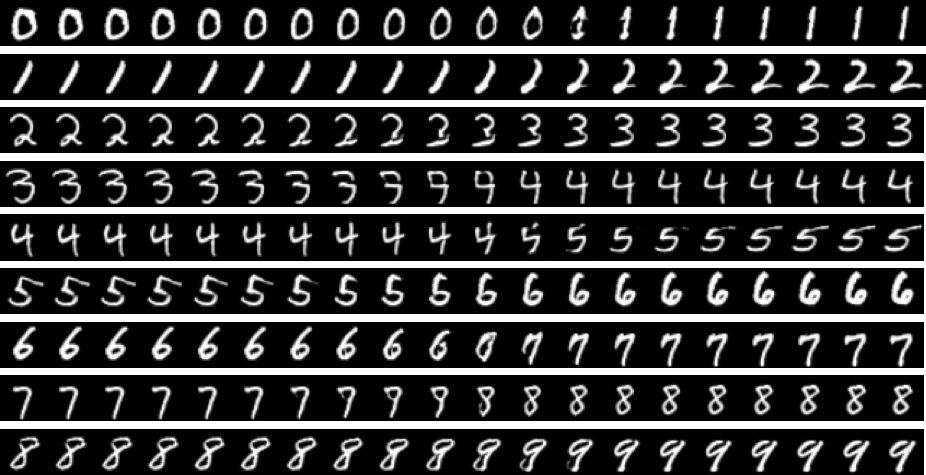} \\
	\end{tabular}
\caption{More examples of digits translations on the MNIST dataset~\cite{lecun1998gradient}. 
    }
    \label{fig:more_mnist}
\end{figure*}

\end{document}